\documentclass{article}

\usepackage[preprint]{neurips_2025}
\usepackage[utf8]{inputenc} 
\usepackage[T1]{fontenc}    
\usepackage{hyperref}       

\usepackage{url}            
\usepackage{booktabs}       
\usepackage{amsfonts}       
\usepackage{nicefrac}       
\usepackage{microtype}      
\usepackage{xcolor}         

\usepackage{graphicx}
\usepackage{caption}
\usepackage{amsmath}
\usepackage{multirow}
\usepackage{colortbl}
\usepackage{makecell}
\usepackage{arydshln}
\usepackage{listings}
\usepackage{afterpage}
\usepackage[ruled]{algorithm2e}
\usepackage{natbib}
\usepackage{subcaption}
\usepackage[most]{tcolorbox}

\definecolor{someorange}{rgb}{0.773,0.353,0.067}
\definecolor{someblue}{rgb}{0.27, 0.35, 0.760}

\title{{\raisebox{-0.15\height}{\includegraphics[width=0.05\textwidth]{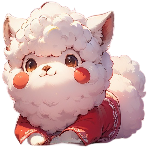}}Continual-NExT: A Unified Comprehension And Generation Continual Learning Framework}}

\author{%
    Jingyang Qiao$^{1,2}$, \ \ Zhizhong Zhang$^{1}$\thanks{Corresponding Author}, \ \ Xin Tan$^{1,3}$, \ \ Jingyu Gong$^{1}$, \\ \textbf{Yanyun Qu}$^{4}$\footnotemark[1], \ \ \textbf{Yuan Xie}$^{1,2}$\thanks{Project Leader} \\
  1. East China Normal University \ \ 2. Shanghai Innovation Institute \\ 3. Shanghai Artificial Intelligence Laboratory \ \ 4. Xiamen University \\
  \url{https://github.com/JingyangQiao/MAGE} \\
  \texttt{52275901010}@stu.ecnu.edu.cn
}

\begin{document}

\maketitle

\begin{abstract} 
Dual-to-Dual MLLMs refer to Multimodal Large Language Models, which can enable unified multimodal comprehension and generation through text and image modalities. Although exhibiting strong instantaneous learning and generalization capabilities, Dual-to-Dual MLLMs still remain deficient in lifelong evolution, significantly affecting continual adaptation to dynamic real-world scenarios. One of the challenges is that learning new tasks inevitably destroys the learned knowledge. Beyond traditional catastrophic forgetting, Dual-to-Dual MLLMs face other challenges, including hallucination, instruction unfollowing, and failures in cross-modal knowledge transfer. However, no standardized continual learning framework for Dual-to-Dual MLLMs has been established yet, leaving these challenges unexplored. Thus, in this paper, we establish \textbf{Continual-NExT}, a continual learning framework for Dual-to-Dual MLLMs with deliberately-architected evaluation metrics. To improve the continual learning capability of Dual-to-Dual MLLMs, we propose an efficient \textbf{MAGE} (\underline{\textbf{M}}ixture and \underline{\textbf{A}}ggregation of \underline{\textbf{G}}eneral LoRA and \underline{\textbf{E}}xpert LoRA) method to further facilitate knowledge transfer across modalities and mitigate forgetting. Extensive experiments demonstrate that MAGE outperforms other continual learning methods and achieves state-of-the-art performance.
\end{abstract}

\section{Introduction} 
\begin{figure*}[t]
    \centering
    \includegraphics[width=0.9\textwidth]{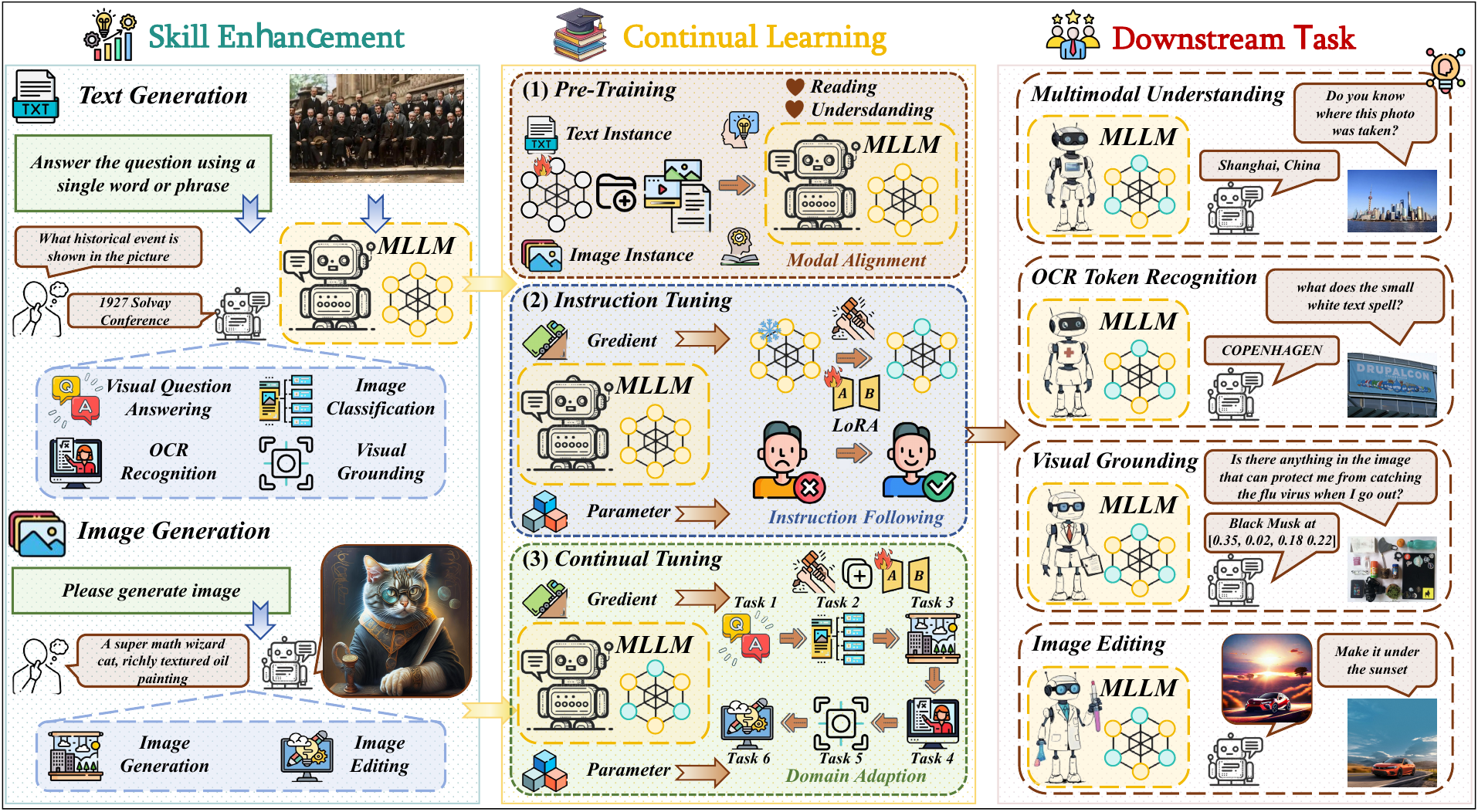}
    \caption{Continual-NExT: A framework for lifelong multimodal learning in Dual-to-Dual MLLMs. The left part shows its multimodal generation capability. The middle part illustrates its development across diverse training stages. The right part presents its supported downstream tasks.}
    \vspace{-5mm}
    \label{fig_1}
\end{figure*}

Recent Multimodal Large Language Models (MLLMs) \citep{instructblipx,llavanext} have gradually evolved from visual question answering to multimodal comprehension and generation \citep{emu}. Dual-to-Dual MLLMs can accept inputs in text/image modality and deliver responses in appropriate text/image feedback \citep{seed}. However, real-world scenarios are characterized by continuously evolving tasks and shifting data distributions. Thus, confronted with these dynamic scenarios, a one-time training paradigm is insufficient. Enhancing the continual learning capability of Dual-to-Dual MLLMs, therefore, becomes increasingly critical \cite{hal,unf,tra}.

Traditional continual learning frameworks for MLLMs predominantly concentrate on improving the text generation capabilities of MLLMs \citep{citb,coin,mcit}, with limited exploration of image and text-image hybrid generation. While continual learning has been extensively studied in text-only generation settings, extending it to Dual-to-Dual MLLMs introduces a less explored but fundamental question: \textit{\textbf{Do multimodal comprehension and generation inherently conflict under continual learning}}\textbf{?}

To answer the question, in this paper, we propose \textbf{Continual-NExT}, the first unified comprehension and generation continual learning framework, as shown in Figure \ref{fig_1}. Continual-NExT encompasses a variety of multimodal input-output tasks, making it more aligned with real-world scenarios and enabling evaluation of Dual-to-Dual MLLMs' capability to continually adapt to a wide range of comprehension and generation tasks.

Based on this framework, we investigate catastrophic forgetting and associated challenges \citep{fgt1,fgt2}. A common intuition is that comprehension and generation represent fundamentally different learning objectives, potentially requiring separate continual learning strategies. However, we discover that the challenge does not stem from the task type itself, but from the asymmetric requirements imposed by input and output modalities. In particular, comprehension tasks primarily rely on stable multimodal representations at the input side, whereas generation tasks demand higher adaptability in modality-specific decoding and reasoning components. In continual learning, these asymmetric requirements lead to distinct forgetting behaviors, even within a single unified model. This observation suggests that the tension between comprehension and generation is not intrinsic, but rather emerges from how model parameters are updated in response to modality-specific inputs and outputs over time.

Therefore, we propose a novel continual learning method: MAGE (\underline{\textbf{M}}ixture and \underline{\textbf{A}}ggregation of \underline{\textbf{G}}eneral LoRA and \underline{\textbf{E}}xpert LoRA). MAGE efficiently inserts trainable LoRA into the frozen LLM and adopts a well-designed collaborative strategy to merge distinct LoRA experts. Among them, General LoRA is responsible for understanding input modalities, and Expert LoRA handles the reasoning of output modalities. Experimental results demonstrate that MAGE significantly enhances the continual learning capabilities of Dual-to-Dual MLLMs. Our contributions are as follows.

\textbullet{} We first present a unified comprehension and generation framework (UCL) with well-defined evaluation metrics, serving to evaluate continual learning performance of Dual-to-Dual MLLMs.

\textbullet{} We introduce an efficient MAGE method that seamlessly integrates General LoRA and Expert LoRA to enhance UCL performance.

\textbullet{} Extensive experiments demonstrate the efficiency of MAGE on the Continual-NExT framework that outperforms other state-of-the-art continual learning methods.

\section{Related Work}
\textbf{Dual-to-Dual MLLMs:} Conventional MLLMs, such as BLIP-2 \citep{blip2}, and Qwen-VL \citep{qwen2vl}, only grasp multimodal comprehension abilities, while lacking multimodal generation abilities. To address this limitation, SEED-X introduces an image discretization tokenizer that encodes visual inputs into discrete tokens with dynamic resolution, facilitating multiscale image understanding and generation \citep{seedx}. NExT-GPT integrates multimodal adapters with various diffusion decoders, allowing it to process inputs and produce outputs across text and image modalities \citep{nextgpt}. AnyGPT employs specialized tokenizers to condense multimodal data, enabling a single autoregressive LLM to handle diverse tasks at the semantic level \citep{anygpt}.

\textbf{Continual Learning:} Traditional continual learning focuses on task-, class-, and domain-incremental learning \citep{trade1}. Recently, continual instruction tuning frameworks have been proposed \citep{citb,coin} and catastrophic forgetting alleviation methods for MLLMs have been explored, including FwT-Prompt employing gradient projection \citep{fwt}, and CIA implementing dynamic updates \citep{cia}. The above frameworks and methods are exclusively based on Dual-to-Text MLLMs. However, continual learning frameworks and methods specifically designed for Dual-to-Dual MLLMs are still blank.

\textbf{Dual-to-Dual MLLMs Continual Learning}: UCL is defined to continually tune Dual-to-Dual MLLMs on new tasks without costly re-training. Compared with traditional continual learning, UCL differs in that it has multimodal inputs and outputs, along with requirements to handle a range of perception and generation tasks. In UCL framework, each incremental task is an independent dataset and focuses on a specific ability. UCL is described as a sequence of tasks $\mathcal{T}_\text{seq} = \{t_1, ..., t_T\}$. Each task $t_j \in \mathcal{T}_\text{seq}$ consists of a training set $\mathcal{S}^{t_j}_{train}$ and a test set $\mathcal{S}^{t_j}_{test}$. The primary goal of UCL is to sequentially learn from $\mathcal{T}_\text{seq}$ while maintaining the balance between plasticity (adaptation to new tasks) and stability (preservation of prior knowledge).

\section{Method} 
\begin{figure}[htb]
\centering
\includegraphics[width=5in]{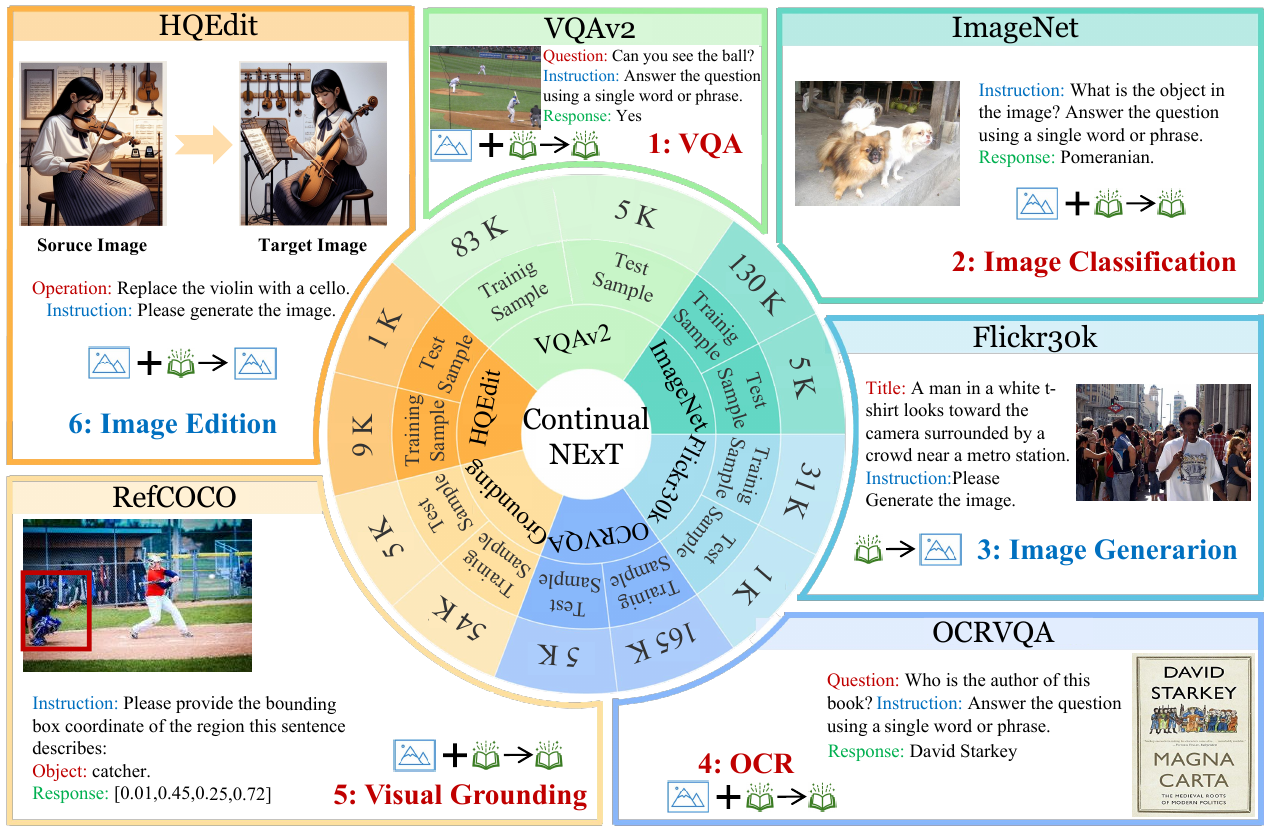}
\caption{Overview of Continual-NExT, including tasks, task types, sizes, and examples.}
\label{fig_2}
\vspace{-.3cm}
\end{figure}

\subsection{Task Settings in Continual-NExT}
Continual-NExT includes six tasks: Visual Question Answering \citep{vqa}, Image Classification \citep{imn}, Image Generation \citep{flickr30k}, OCR Token Recognition \citep{oqa}, Visual Grounding \citep{refcoco1,refcoco2}, and Image Editing \citep{hq}. Their diversity across modalities and domains ensures a robust evaluation of multimodal comprehension and generation capabilities. We construct the instructions of text generation tasks (\textit{e.g.}, VQAv2, Imagenet-1K, OCRVQA, and RefCOCO) by following \citep{coin}. Additionally, we adopt the self-made instructions in image generation tasks (\textit{e.g.}, Flickr30K and HQEdit). Instructions, data, and statistical details of each task are illustrated in Figure \ref{fig_2}. The amount of training data in each task is varied, ranging from 10K to 100K, which simulates the randomness of data distribution in the real world. (Detailed information of each task is in Appendix \ref{datasetapp}).

\subsection{Advanced Evaluation Scenarios}
We utilize popular evaluation protocols \citep{cia}, including Average Accuracy (Avg.Acc), Forgetting, and New Accuracy (New.Acc). For evaluating on text generation tasks, Accuracy is calculated by directly comparing the predictions with labels \citep{coin}. For evaluating on image generation tasks, Accuracy is calculated by employing the CLIP Score \citep{cs}. We discover that forgetting in UCL is mainly caused by hallucination, instruction unfollowing, and the other errors. Thus, we introduce three diagnostic evaluation metrics to better analyze forgetting behaviors. Specifically, we define the Average Hallucination Rate (Avg.HAL) to depict hallucination-related forgetting, the Average Instruction-Unfollowing Rate (Avg.IUF) to show instruction unfollowing-related forgetting, and the Average Other Error Rate (Avg.OTH) to summarize the other errors-related forgetting (Metrics calculation can be found in Appendix \ref{metricapp}).

\textbf{\textit{Discussion}:} Existing continual learning frameworks for MLLMs mainly rely on engineering-driven unification of data or instruction templates, providing limited insight into the mechanisms of forgetting across heterogeneous tasks. In contrast, Continual-NExT is not merely a collection of multimodal tasks, but a framework designed to expose and evaluate how different input-output modality combinations induce distinct patterns of forgetting and transfer within a unified model.

\section{MAGE: An Efficient UCL Framework}
\subsection{Motivation}
\begin{figure}[htb]
    \centering
    \includegraphics[width=0.9\textwidth]{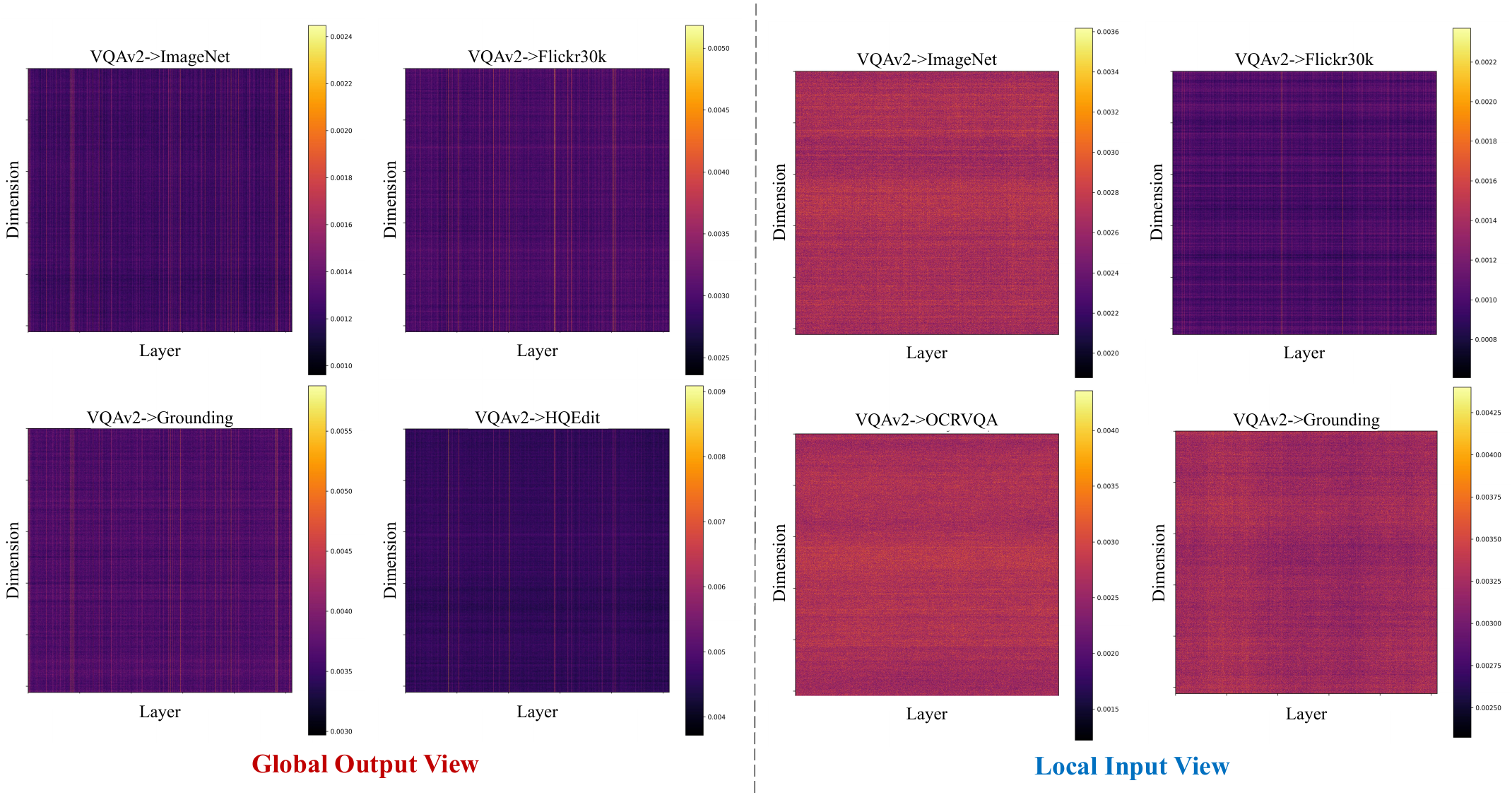}
    \caption{Parameter update patterns across heterogeneous tasks from global output view (left) to local input view (right). VQAv2->ImageNet refers to first training on VQAv2 (Task 1) and then training on ImageNet (Task 2). Pixel value represents the absolute discrepancy between the parameters learned from Task 1 and Task 2.}
    \vspace{-3mm}
    \label{fig_3}
\end{figure}

Existing research shows that the model focuses on different input patterns at each layer, indicating that the model updates its parameters in response to task-specific inputs \citep{task}. However, it is unknown \textit{\textbf{whether the model updates its parameters in response to both task-specific inputs and outputs.}} To answer the question, we fine-tune SEED-X \cite{seedx} using LoRA and analyze the parameter update patterns across tasks, as shown in Figure \ref{fig_3}. We select VQAv2 as the first task, followed by five distinct tasks as the second task, including ImageNet, Grounding, OCRVQA (image and text as inputs, text as outputs), Flickr30k (text as inputs, image as outputs), and HQEdit (image and text as inputs, image as outputs). We visualize the absolute discrepancy between the parameters learned from Task 1 and Task 2 and depict heat maps.

For input-related patterns, we analyze the absolute discrepancy in the shallow (the former 50\%) layers according to \citep{shallow}. Tasks using both text and image inputs (\textit{e.g.}, ImageNet, Grounding, OCRVQA) show high discrepancies across most parameters, indicating the need for both text and image comprehension. In contrast, Flickr30k shows high discrepancies in fewer parameters, as it primarily involves text comprehension. For output-related patterns, we examine the absolute discrepancy in the deep (the last 50\%) layers according to \citep{whole}. Text-output tasks (\textit{e.g.}, ImageNet, Grounding) concentrate their discrepancies in a small set of parameters primarily associated with text-generation, whereas image-output tasks (\textit{e.g.}, Flickr30k, HQEdit) exhibit discrepancies in another different and small set of parameters tied to image-generation. Notably, the overlap between text- and image-output related parameters remains isolated.

\begin{figure}[htb]
    \centering
    \includegraphics[width=0.6\textwidth]{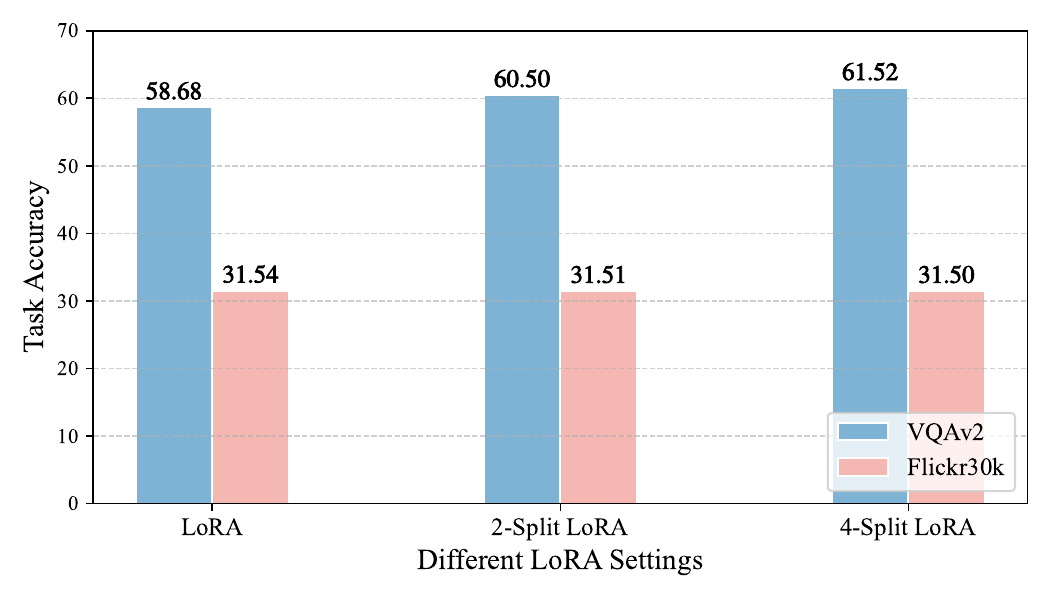}
    \vspace{-3mm}
    \caption{Results of different LoRA settings.}
    \vspace{-3mm}
    \label{pre}
\end{figure}

Motivated by the discovery, we propose two types of LoRA: General LoRA for modality comprehension and Expert LoRA for modality generation. General LoRA includes Image General LoRA for understanding images and Text General LoRA for understanding text. Similarly, Expert LoRA includes Image Expert LoRA for image generation and Text Expert LoRA for text generation. Considering that fine-tuning all LoRAs may lead to severe catastrophic forgetting, and the model primarily focuses on a particular subset of parameters in the specific task, we propose fine-tuning only the LoRA related to the task's input and output modalities while freezing others. This approach enables learning the new task while preserving existing knowledge. For validation, we first fine-tune the LoRA on VQAv2 and then on Flickr30k. We adopt three settings: (\textit{i}) LoRA, (\textit{ii}) 2-Split LoRA splitting into two parts and updating according to the task-specific input modality, (\textit{iii}) 4-Split LoRA splitting into four parts and updating according to both the task-specific input and output modalities. The total rank of the three types of LoRA is the same. As shown in Figure \ref{pre}, the proposed method (\textit{iii}) can achieve the best anti-forgetting ability with fewer reductions in new task learning performance.

\begin{figure*}[thb]
\centering
\includegraphics[width=5in]{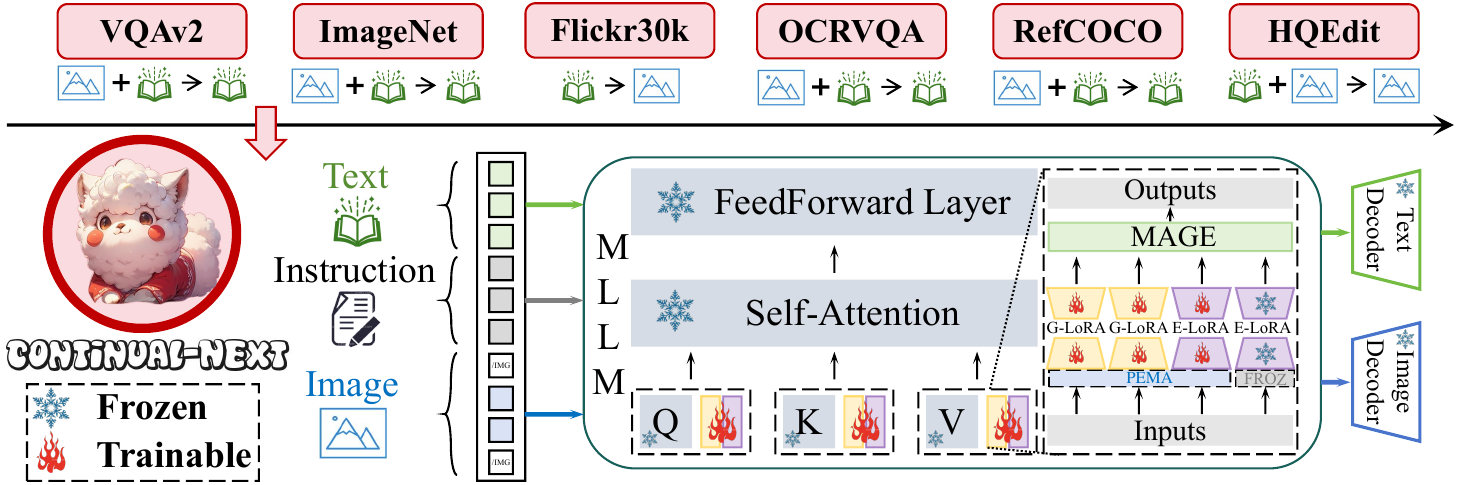}
\caption{Overview of the MAGE. We insert four types of LoRA into the MLLM, Image/Text General LoRA (G-LoRA) and Image/Text Expert LoRA (E-LoRA). Then, we use the parameter-wise EMA strategy to update LoRA corresponding to the specific input and output modality of current task, while freezing that unrelated to it.}
\label{fig_4}
\vspace{-.3cm}
\end{figure*}

Furthermore, we integrate General LoRA and Expert LoRA through an equal-weight linear summation as in Eq.(\ref{mage}). $\mathcal{W}_o$ means original parameters in LLM, $\mathcal{W}_{down}^{GI}$ and $\mathcal{W}_{up}^{GI}$ denote General Image LoRA parameters, $\mathcal{W}_{down}^{GT}$ and $\mathcal{W}_{up}^{GT}$ represent General Text LoRA parameters, $\mathcal{W}_{down}^{EI}$ and $\mathcal{W}_{up}^{EI}$ indicate Expert Image LoRA parameters, $\mathcal{W}_{down}^{ET}$ and $\mathcal{W}_{up}^{ET}$ refer to Expert Text LoRA parameters. $x$ and $h$ are inputs and outputs.
\begin{align}
\label{mage}
h = (\mathcal{W}_o 
       + \mathcal{W}_{down}^{GI}\mathcal{W}_{up}^{GI} 
       + \mathcal{W}_{down}^{GT}\mathcal{W}_{up}^{GT}
       + \mathcal{W}_{down}^{EI}\mathcal{W}_{up}^{EI} 
       + \mathcal{W}_{down}^{ET}\mathcal{W}_{up}^{ET}) \, x.
\end{align}

Notably, although the frozen LoRA still participates in the prediction, it remains inactive and therefore does not influence the output. This design ensures that trainable LoRA facilitates cross-modal knowledge transfer while frozen LoRA prevents forgetting of modality-specific knowledge.

\textbf{\textit{Discussion}:} We analyze parameter update patterns across tasks with varying input–output modalities. Based on this insight, we propose MAGE, which explicitly separates modality comprehension components from modality generation components via structured LoRA specialization. This empirical observation serves as a motivating insight rather than a definitive conclusion. It highlights a practical direction for structuring parameter-efficient adaptations in continual learning.

\subsection{Mechanism of Anti-Forgetting} 
To further mitigate forgetting, we draw inspiration from the Dynamic Exponential Moving Average (DEMA) method proposed by \citep{cia}. However, its original setting is limited in 1) requiring a differencing-based approximation of the Hessian matrix, leading to introduce additional memory overhead (\textit{e.g.}, storing historical parameters and gradients), 2) relying on regularization to calculate EMA weights, resulting in a coarse, layer-wise granularity. To address these limitations, we revisit the derivation of dynamic EMA and propose a novel PEMA (Parameter-wise EMA) method.

Inspired by \citep{ewc}, we utilize the Fisher matrix to replace with Hessian matrix because the Fisher matrix equals the negative expected Hessian matrix. Leveraging Monte Carlo sampling to approximate this expectation, we compute the Fisher matrix by averaging the gradients of each sample in one mini-batch as shown in:
\begin{equation}
\label{fisher}
H \approx F_{ii} = \frac{1}{|B|} \sum_{x \in B} 
\left( 
    \left. \frac{\partial L (y, f(x))}{\partial \theta} \right|_{\theta = \theta_t}
\right)^2,
\end{equation}
where $F$ is the Fisher matrix, $B$ denotes one mini-batch, $L$ refers to the loss function, $(x,y)$ means the input and label, $f$ represents the model and $\theta_t$ indicates the parameters trained after task $t$.

The calculation of dynamic EMA weight requires the inverse Hessian. Since the Hessian matrix is not always invertible, we adopt a regularized inverse way, as shown in Eq.(\ref{regrev}). We add a term $\lambda I$ to the Hessian matrix, where $\lambda$ is a hyperparameter (set to 1e-5) and $I$ denotes the identity matrix. This procedure is equivalent to applying “damping” to the Hessian matrix, making nearly singular directions invertible and preventing numerically unstable exploding inverses, thereby ensuring that the modified Hessian matrix $\hat{H}$ is always invertible.
\begin{equation}
\label{regrev}
\hat{H} = H^TH + \lambda I.
\end{equation}

Finally, we obtain the parameter-wise EMA weights through matrix multiplication as in Eq.(\ref{weight}).
\begin{equation}
\label{weight}
\beta_t = [\frac{\partial L}{\partial \theta_t} + I][(\theta_t-\theta_{t-1}^*)^T\hat{H}]^{-1},
\end{equation}
where $\theta_{t-1}^*$ is the EMA parameters trained after task $t-1$. We utilize the PEMA method to update LoRA corresponding to the specific input and output modality of the current task. Detailed deduction process is in Appendix \ref{deduction}. Comparisons between PEMA and DEMA are in Appendix \ref{addiexp}.

\section{Experiments}
\begin{table*}[th]
\vspace{-3mm}
\renewcommand{\arraystretch}{2.0}
\caption{Avg.ACC, Forgetting, and New.ACC performance comparisons. Accuracy on each task refers to the performance after training on the current(first row)/final(second row) task. \textbf{Bold} fonts represent the best performance, and \underline{underlined} fonts represent the second-best performance. Upper-Bound denotes once training all the datasets. \looseness=-1}
\label{compare}
\centering
\resizebox{1.0\linewidth}{!}{
\begin{tabular}{l|c|cccccc|ccc}
\toprule
\multirow{2}{*}{\textbf{Method}} & \multirow{2}{*}{\textbf{Venue}} & \multicolumn{6}{c|}{\textbf{Accuracy on Each Task}} & \multicolumn{3}{c}{\textbf{Overall Results}} \\ \cline{3-11} 
& & \multicolumn{1}{c}{\textbf{\textcolor{someblue}{VQAv2}}} & \multicolumn{1}{c}{\textbf{\textcolor{someblue}{ImageNet}}} & \multicolumn{1}{c}{\textbf{\textcolor{someblue}{Flickr30k}}} & \multicolumn{1}{c}{\textbf{\textcolor{someblue}{OCRVQA}}} & \multicolumn{1}{c}{\textbf{\textcolor{someblue}{RefCOCO}}} & \multicolumn{1}{c|}{\textbf{\textcolor{someblue}{HQEdit}}} & \multicolumn{1}{c}{\textbf{\textcolor[rgb]{0.773,0.353,0.067}{Avg.ACC($\uparrow$)}}} & \multicolumn{1}{c}{\textbf{\textcolor[rgb]{0.773,0.353,0.067}{Forgetting($\downarrow$)}}} & \multicolumn{1}{c}{\textbf{\textcolor[rgb]{0.773,0.353,0.067}{New.ACC($\uparrow$)}}} \\ \hline
Zero-Shot & - & 26.64 & 28.38 & 19.76 & 13.78 & 25.54 & 58.76 & 28.81 & - & - \\
\hdashline
\multirow{2}{*}{LoRA \citep{lora}} & \multirow{2}{*}{ICLR'22} & 59.68 & 65.19 & 31.18 & 41.22 & 66.60 & 91.27 & \multirow{2}{*}{42.23} & \multirow{2}{*}{20.36} & \multirow{2}{*}{59.19} \\
 &  & 18.40 & 61.01 & 30.60 & 19.38 & 32.70 & 91.27 & & & \\
\hdashline
\multirow{2}{*}{MoELoRA \citep{coin}} & \multirow{2}{*}{NIPS'24} & 62.10 & 82.16 & 31.46 & 21.52 & 70.00 & 91.54 & \multirow{2}{*}{43.90} & \multirow{2}{*}{19.08} & \multirow{2}{*}{59.80} \\
 &  &  25.32 & 63.57 & 30.59 & 16.04 & 36.32 & 91.54 &  &  &  \\
\hdashline
\multirow{2}{*}{EWC \citep{ewc}} & \multirow{2}{*}{PNAS'17} & 59.68 & 73.29	& 31.11	& 37.50	& 68.40 & 91.30 & \multirow{2}{*}{47.00} & \multirow{2}{*}{15.85} & \multirow{2}{*}{\underline{60.21}} \\
 &  &  40.44 & 61.07 & 30.49 & 18.64 & 40.08 & 91.30  &  &  &  \\
\hdashline
\multirow{2}{*}{LAE \citep{lae}} & \multirow{2}{*}{ICCV'23} & 59.68 & 77.49 & 31.18 & 27.48 & 65.72 & 91.39 & \multirow{2}{*}{45.25} & \multirow{2}{*}{16.29} & \multirow{2}{*}{58.82} \\
 &  & 21.56 & 70.71 & 30.63 & 28.54 & 28.68 & 91.39 &  &  &  \\
\hdashline
\multirow{2}{*}{PGP \citep{pgp}} & \multirow{2}{*}{ICLR'24} & 59.68	& 78.20	& 31.40	& 19.96	& 73.20	& 91.43 & \multirow{2}{*}{47.70} & \multirow{2}{*}{\underline{13.54}} & \multirow{2}{*}{58.97} \\
 &  & 38.08 & 74.14 & 30.82 & 18.08 & 33.64 & 91.43 &  &  &  \\
\hdashline
\multirow{2}{*}{CIA \citep{cia}} & \multirow{2}{*}{ICML'25} & 59.68	& 68.95	& 31.08	& 41.06	& 70.28	& 91.36 & \multirow{2}{*}{\underline{47.99}} & \multirow{2}{*}{14.89} & \multirow{2}{*}{\textbf{60.40}} \\
 & & 28.40 & 68.91 & 30.49 & 31.00 & 37.78 & 91.36 &  &  &  \\
\hdashline
\multirow{2}{*}{RegLoRA \citep{sefe}} & \multirow{2}{*}{ICML'25} & 59.68 & 73.31 & 31.29 & 36.92 & 66.82 & 91.60 & \multirow{2}{*}{47.39} & \multirow{2}{*}{15.05} & \multirow{2}{*}{59.94} \\
 &  & 40.34 & 61.10 & 30.42 & 25.72 & 35.18 & 91.60 &  &  &  \\
\hdashline
\multirow{2}{*}{\textbf{Ours(MAGE)}} & \multirow{2}{*}{\textbf{-}} & 62.10 & 68.35 & 31.26 & 37.09 & 68.74 & 91.19 & \multirow{2}{*}{\textbf{49.58}} & \multirow{2}{*}{\textbf{12.26}} & \multirow{2}{*}{59.79} \\ 
 &  & 45.42 & 68.02 & 31.44 & 27.28 & 34.10 & 91.19 &  &  &  \\
\hdashline
\color{gray}\textbf{Upper-Bound} & \color{gray}\textbf{-} & \color{gray}\textbf{62.72} & \color{gray}\textbf{47.88}	& \color{gray}\textbf{31.24}	& \color{gray}\textbf{42.84}	& \color{gray}\textbf{47.68}	& \color{gray}\textbf{87.49}	& \color{gray}\textbf{53.31} & \color{gray}\textbf{-} & \color{gray}\textbf{-} \\
\bottomrule
\end{tabular}}
\vspace{-3mm}
\end{table*}

\label{expe}
\subsection{Experimental Setup}
\textbf{Implementation Details:} We adopt SEED-X \citep{seedx} as our backbone with inserted LoRA \citep{lora} in the LLM side, LLaMA2-chat-13B \citep{llama2}. During the unified comprehension and generation continual learning, we freeze the vision encoder and LLM, with only training connection layers and LoRA (Detailed information is listed in Appendix \ref{trainapp}). The continual learning orders are VQAv2 \citep{vqa}, ImageNet \citep{imn}, Flickr30k \citep{flickr30k}, OCRVQA \citep{oqa}, RefCOCO \citep{refcoco1, refcoco2}, and HQEdit \citep{hq}.

\textbf{Compared Methods\footnote{All methods hold an equal scale of trainable parameters.}:} (1) Zero-Shot, (2) LoRA Fine-Tune \citep{lora}, (3) MoELoRA \citep{coin}, (4) LAE \citep{lae}, (5) EWC \citep{ewc}, (6) PGP \citep{pgp}, (7) CIA \citep{cia}, (8) RegLoRA \citep{sefe}. Details are in Appendix \ref{beseapp}. 

\subsection{Compared with baselines}
We compare the performance of our method with baselines in Table \ref{compare}. Under non-model-expansion settings, we observe that MAGE outperforms the best among other methods, CIA by +1.59@Avg.ACC and -2.63@Forgetting. Compared with LoRA Fine-Tune, MoELoRA has superior anti-forgetting performance due to the multi-experts mixture mechanism, while it fails in some tasks. Although methods like LAE and PGP can resist forgetting, their plasticity is influenced. EWC can perform well both in stability and plasticity, while the Avg.ACC needs to be improved. CIA and RegLoRA own both extraordinary new task learning and old task anti-forgetting abilities, while they require saving intermediate variables, leading to high memory consumption. MAGE outperforms all other baselines with less memory usage.

\subsection{Forgetting Analysis}
We perform an in-depth analysis of forgetting in UCL and discover that it is mainly caused by three categories. The first category is instruction unfollowing, which refers to the model's failure to follow input instructions when generating responses. We attribute this issue to semantic discrepancies among the instructions. The second category is hallucination, which means the model can not perform accurate predictions based on existing knowledge, resulting in the production of factually incorrect or illusory content. We consider hallucination to arise from the interference between new and old knowledge during continual task adaptation. Compared to instruction unfollowing, hallucination responses generally align well with the syntactic or structural demands of the instruction, while significantly differing from the ground truth. All other errors that do not fall into the above two categories are classified in the third category, presented as the other errors. Note that cases exhibiting both hallucination and instruction unfollowing are uniformly deemed instruction unfollowing errors.

\begin{figure*}[htb]
\centering
\includegraphics[width=5.5in]{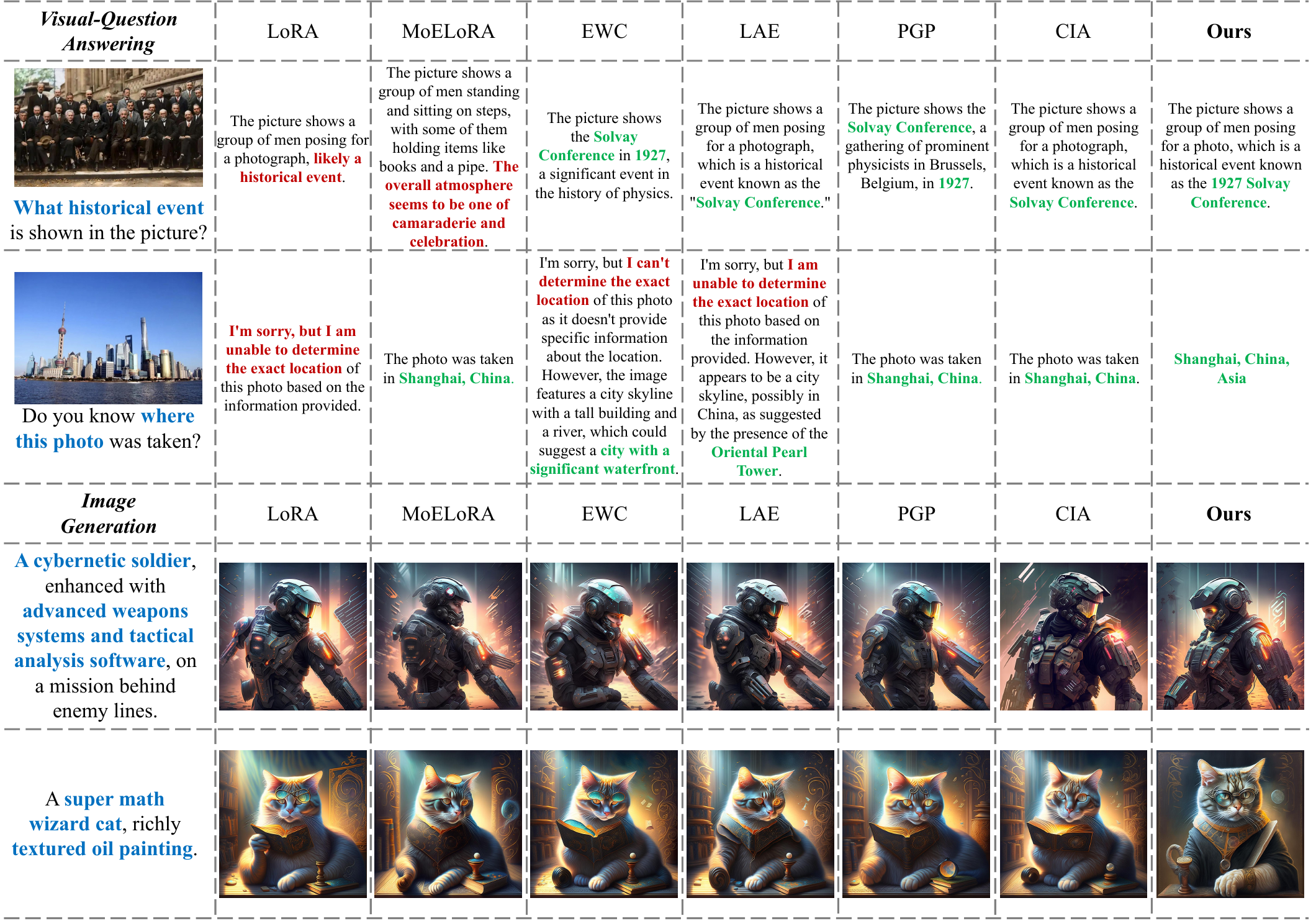}
\caption{Visualized examples for each method after UCL on the Continual-NExT framework.}
\label{fig_6}
\vspace{-.3cm}
\end{figure*}

\begin{table}[h]
\caption{Comparisons on Avg.IUF, Avg.HAL and Avg.OTH between LoRA, MoELoRA, and MAGE.}
\label{mix}
\centering
\scalebox{1.00}{
\begin{tabular}{l|ccc}
\toprule
\multirow{2}{*}{\textbf{Method}} & \multicolumn{3}{c}{\textbf{Forgetting Type Metrics}} \\ \cline{2-4} 
& \textcolor[rgb]{0.773,0.353,0.067}{\textbf{Avg.IUF($\downarrow$)}}   & \multicolumn{1}{c}{\textcolor[rgb]{0.773,0.353,0.067}{\textbf{Avg.HAL($\downarrow$)}}} & \multicolumn{1}{c}{\textcolor[rgb]{0.773,0.353,0.067}{\textbf{Avg.OTH($\downarrow$)}}} \\ \hline
LoRA          & 9.07 & 11.11 & \underline{0.18} \\ \hdashline
MoELoRA       & \underline{8.47} & \underline{10.53} & \textbf{0.08} \\ \hdashline
\textbf{MAGE} & \textbf{4.42} & \textbf{7.48} & 0.36 \\ \bottomrule
\end{tabular}}
\vspace{-3mm}
\end{table}

\begin{table*}[ht]
\caption{Knowledge transfer comparisons between LoRA and MAGE in Continual-NExT.}
\label{kt}
\centering
\renewcommand{\arraystretch}{1.4} 
\resizebox{0.9\linewidth}{!}{
\begin{tabular}{l|c:c|c:c|>{}c:>{}c|>{}c:>{}c}
\toprule
\textbf{Method} & \multicolumn{4}{c|}{\textbf{LoRA}} & \multicolumn{4}{c}{\textbf{MAGE}} \\ \hline
\textbf{Tra.Task} & \textbf{\textcolor{someblue}{Flickr30k}} & \textbf{\textcolor[rgb]{0.773,0.353,0.067}{OCRVQA}}  & \textbf{\textcolor{someblue}{Flickr30k}} & \textbf{\textcolor[rgb]{0.773,0.353,0.067}{OCRVQA}} & \textbf{\textcolor{someblue}{Flickr30k}} & \textbf{\textcolor[rgb]{0.773,0.353,0.067}{OCRVQA}} & \textbf{\textcolor{someblue}{Flickr30k}} & \textbf{\textcolor[rgb]{0.773,0.353,0.067}{OCRVQA}} \\ \hline
\textbf{Evl.Task} & VQAv2 & VQAv2 & Flickr30k & Flickr30k & VQAv2 & VQAv2 & Flickr30k & Flickr30k \\ \hline
\textbf{Trans.Type} & \multicolumn{2}{c|}{Text$\rightarrow$Text} & \multicolumn{2}{c|}{Text$\rightarrow$Image} & \multicolumn{2}{c|}{ Text$\rightarrow$Text} & \multicolumn{2}{c}{ Text$\rightarrow$Image} \\ \hline
\textbf{Accuracy} & 20.10 & 50.26 & 31.18 & 31.26 & 43.56 & 57.29 & 31.26 & 32.02 \\ \bottomrule
\end{tabular}}
\renewcommand{\arraystretch}{1.0} 
\vspace{-1mm}
\end{table*}

We compare the performance of MAGE, MoELoRA, and LoRA Fine-Tune under Avg.HAL, Avg.IUF and Avg.OTH metrics in Table \ref{mix}. With the multi-experts mixture mechanism, MoELoRA alleviates overall forgetting compared to LoRA Fine-Tune. Our method achieves notable reductions in Avg.HAL and Avg.IUF, while slight increases in Avg.OTH. This can be attributed to errors originally caused by hallucination and instruction unfollowing being transferred to the other error. These experiments provide explanations for the superior anti-forgetting ability of MAGE.

\begin{table*}[h]
\renewcommand{\arraystretch}{1.5}
\caption{Ablation study on different method components and their influence on each task's performance. The best and second-best results are marked in \textbf{bold} and \underline{underlined}, respectively. \looseness=-1}
\label{ab1}
\centering
\huge
\resizebox{0.9\linewidth}{!}{
\begin{tabular}{l|llllll|lll}
\toprule
\multirow{2}{*}{\textbf{Method}} & \multicolumn{6}{c|}{\textbf{Accuracy on Each Task}} & \multicolumn{3}{c}{\textbf{Overall Results}} \\ \cline{2-10} 
& \multicolumn{1}{c}{\textbf{\textcolor{someblue}{VQAv2}}} & \multicolumn{1}{c}{\textbf{\textcolor{someblue}{ImageNet}}} & \multicolumn{1}{c}{\textbf{\textcolor{someblue}{Flickr30k}}} & \multicolumn{1}{c}{\textbf{\textcolor{someblue}{OCRVQA}}} & \multicolumn{1}{c}{\textbf{\textcolor{someblue}{RefCOCO}}} & \multicolumn{1}{c|}{\textbf{\textcolor{someblue}{HQEdit}}} & \multicolumn{1}{c}{\textbf{\textcolor[rgb]{0.773,0.353,0.067}{Avg.ACC($\uparrow$)}}} & \multicolumn{1}{c}{\textbf{\textcolor[rgb]{0.773,0.353,0.067}{Forgetting($\downarrow$)}}} & \multicolumn{1}{c}{\textbf{\textcolor[rgb]{0.773,0.353,0.067}{New.ACC($\uparrow$)}}} \\ \hline
MoE & \multicolumn{1}{c}{25.32} & \multicolumn{1}{c}{\underline{63.57}} & \multicolumn{1}{c}{30.59} & \multicolumn{1}{c}{16.04} & \multicolumn{1}{c}{\textbf{36.32}} & \multicolumn{1}{c|}{\textbf{91.54}} & \multicolumn{1}{c}{43.90} & \multicolumn{1}{c}{19.08} & \multicolumn{1}{c}{59.80} \\ \hdashline
EW & \multicolumn{1}{c}{\underline{47.28}} & \multicolumn{1}{c}{60.93} & \multicolumn{1}{c}{\underline{31.25}} & \multicolumn{1}{c}{18.46} & \multicolumn{1}{c}{31.30} & \multicolumn{1}{c|}{\textbf{91.29}} & \multicolumn{1}{c}{46.75} & \multicolumn{1}{c}{15.66} & \multicolumn{1}{c}{\underline{59.81}} \\ \hdashline
EW+GA & \multicolumn{1}{c}{\textbf{48.54}} & \multicolumn{1}{c}{60.59} & \multicolumn{1}{c}{30.97} & \multicolumn{1}{c}{\underline{20.76}} & \multicolumn{1}{c}{33.92} & \multicolumn{1}{c|}{91.25} & \multicolumn{1}{c}{\underline{47.67}} & \multicolumn{1}{c}{\underline{14.90}} & \multicolumn{1}{c}{\textbf{60.09}} \\ \hdashline
\textbf{MAGE} & \multicolumn{1}{c}{45.42} & \multicolumn{1}{c}{\textbf{68.02}} & \multicolumn{1}{c}{\textbf{31.44}} & \multicolumn{1}{c}{\textbf{27.28}} & \multicolumn{1}{c}{\underline{34.10}} & \multicolumn{1}{c|}{91.19} & \multicolumn{1}{c}{\textbf{49.58}} & \multicolumn{1}{c}{\textbf{12.26}} & \multicolumn{1}{c}{59.79} \\ \bottomrule
\end{tabular}}
\vspace{-3mm}
\end{table*}

\subsection{Knowledge Transfer}
We investigate the knowledge transfer phenomenon in UCL, as shown in Table \ref{kt}. We primarily analyze intra-modal knowledge transfer, which occurs between tasks sharing similar instruction templates. To verify the effect of knowledge transfer in text modality, we evaluate performance on VQAv2 (Task 1) by using models trained on Flickr30k (Task 3) and OCRVQA (Task 4), respectively. In LoRA Fine-Tune, we observe that after training on Flickr30k, MLLM retains 20.10 accuracy on VQAv2, indicating a severe loss of previously learned knowledge. However, after training on OCRVQA, Accuracy on VQAv2 significantly improves to 50.26. This can be attributed to the knowledge transfer between OCRVQA and VQAv2, allowing the model to “recall” what it had previously learned in VQAv2. In MAGE, we witness a similar phenomenon after training on OCRVQA, the model’s performance on VQAv2 also has a substantial improvement. Notably, after training on Flickr30k, MAGE maintains a much higher Accuracy on VQAv2 (43.56) compared to LoRA Fine-Tune (20.10), which demonstrates the strong anti-forgetting capability of our method. 

We further explore cross-modal knowledge transfer, which appears between tasks with similar semantics across modalities. We evaluate on Flickr30k by using models trained on Flickr30k and OCRVQA. Results confirm that training on OCRVQA further improves MLLMs' performance on Flickr30k rather than degrading on this task, suggesting efficient text-to-image knowledge transfer. These findings resonate with some novel viewpoints in continual learning, like MLLMs do not really forget previous knowledge, and previous task capabilities could be restored through adaptation to new tasks. The observed knowledge transfer phenomena may provide experimental support and research perspectives for these opinions.

\subsection{Robust Performance}
To validate the robustness of MAGE, we adopt three training orders (Detailed information is in Appendix \ref{orderapp}). In Table \ref{order}, we find that New.ACC and Avg.ACC have an observed range of fluctuations. Forgetting, as an important performance metric, has a small range of variation (from 12.17 to 12.35). Based on the observations, we conclude that the fluctuations in New.ACC and Avg.ACC are caused by changes in the task training order. However, our method has strong robustness, maintaining the Forgetting metric at a stable level.

\begin{table}[h]
\vspace{-3mm}
\caption{UCL performance under different task orders.}
\label{order}
\centering
\renewcommand{\arraystretch}{1.4} 
\scalebox{0.40}{
\huge{
\begin{tabular}{l|cc|ccc}
\toprule
\multirow{2}{*}{\textbf{Method}} & \multicolumn{2}{c|}{\textbf{Modality}} & \multicolumn{3}{c}{\textbf{Overall Results}} \\ \cline{2-6} 
& \textbf{\textcolor{someblue}{Text}} & \multicolumn{1}{c|}{\textbf{\textcolor{someblue}{Image}}} & \textbf{\textcolor[rgb]{0.773,0.353,0.067}{Avg.ACC($\uparrow$)}} & \multicolumn{1}{c}{\textbf{\textcolor[rgb]{0.773,0.353,0.067}{Forgetting($\downarrow$)}}} & \multicolumn{1}{c}{\textbf{\textcolor[rgb]{0.773,0.353,0.067}{New.ACC($\uparrow$)}}} \\ \hline
Default & 43.71 & \multicolumn{1}{c|}{61.32} & 49.58 & \multicolumn{1}{c}{12.26} & \multicolumn{1}{c}{59.79} \\ \hdashline
Reverse & \multicolumn{1}{c}{46.07} & \multicolumn{1}{c|}{56.45} & \multicolumn{1}{c}{49.53} & \multicolumn{1}{c}{12.17} & \multicolumn{1}{c}{59.67} \\ \hdashline
Alphabet & \multicolumn{1}{c}{45.18} & 57.31 & \multicolumn{1}{c}{49.22} & \multicolumn{1}{c}{12.35} & \multicolumn{1}{c}{59.51} \\ \bottomrule
\end{tabular}}}
\renewcommand{\arraystretch}{1.0} 
\vspace{-3mm}
\end{table}


\subsection{Ablation Studies}
We progressively incorporate proposed modules to evaluate their contributions, starting from the simple MoELoRA baseline with four experts. Experimental results are shown in Table \ref{ab1}. First, we replace the routing function in MoELoRA with an equal-weighted summation of outputs from LoRA experts and the base model (denoted as EW). This module increases the Avg.ACC from 43.90 to 46.75 and reduces the Forgetting from 19.08 to 15.66. Then, we introduce the General LoRA and Expert LoRA fusion mechanism (denoted as EW+GA) by increasing the Avg.ACC from 46.75 to 47.67 and decreasing the Forgetting from 15.66 to 14.90. Finally, we adopt the parameter-wise EMA update strategy (denoted as MAGE), which increases the Avg.ACC up to 49.58 and lowers the Forgetting to 12.26. Ablation studies demonstrate that each of the proposed modules contributes independently and significantly to the overall performance.

\subsection{Descriptive Statistics}
A full continual learning and evaluation pipeline spanning six tasks takes about one week to complete by using 8 NVIDIA H100 GPUs. We report the descriptive statistics about our results in Appendix \ref{addiexp} under distinct random seed settings.

\section{Conclusion}
To evaluate the continual learning ability of Dual-to-Dual MLLMs, we establish a novel framework, Continual-NExT, with well-defined evaluation metrics. Furthermore, we discover that activated parameters are closely related to input modalities and output modalities of the training task and propose an efficient MAGE method. Additionally, we propose a fine-grained and parameter-wise EMA method to mitigate forgetting. Experiments show that our method has excellent anti-forgetting ability and continual learning performance. In the future, we aim to extend our framework and method to more modalities and stronger Any-to-Any MLLMs.

\section{Limitations}
For limitations, as the number of modalities is incrementally growing, the corresponding increase in the number of LoRA experts may lead to a linear explosion in both computational cost and memory consumption. Note that both image and text are seen modalities for the adopted MLLM backbone, whether MLLMs can incrementally incorporate unseen modalities remains an open question. Our future research will focus on how to effectively address the above challenges. Additionally, we note that alternative decompositions, such as finer-grained splits, hierarchical structures, or dynamically routed adaptations, are plausible extensions, and we leave their exploration to future work.

\section{Ethical Considerations}
This work proposes a continual learning framework for unified comprehension and generation using publicly available datasets. The proposed method does not introduce new data sources or application scenarios beyond standard benchmarks, and thus does not raise additional ethical concerns beyond those inherent to the underlying pretrained models and datasets. Potential risks include the propagation of biases present in the pretrained models, as well as possible performance degradation in safety-critical or out-of-distribution settings. We do not evaluate our method in high-risk real-world applications, and recommend caution when deploying the framework in such scenarios.

All datasets used in this work are publicly available benchmarks that have been widely used in prior research. We do not collect any new data, and the datasets do not contain personally identifying information beyond what is already present in the original releases. We follow the original dataset licenses and usage guidelines, and do not perform any additional data processing that would introduce privacy risks. As such, our work does not raise new concerns regarding personal data or offensive content beyond those associated with existing benchmarks.

\section{Conclusion}

\bibliographystyle{unsrt}
\bibliography{references_tech}

\clearpage

\section*{Appendix}

\label{sec:appendix}

\subsection{Task and Dataset Details}
\label{datasetapp}
\textbf{VQAv2} \citep{vqa}: The Visual Question Answering v2.0 (VQA v2.0) dataset is a widely used multimodal benchmark. It is designed to evaluate a model's ability to answer open-ended questions about images, requiring an understanding of both visual content and language based upon the MS COCO dataset. Each question is associated with 10 ground truth answers provided by human annotators, capturing the variability in human responses. Each image has an average of 5.4 questions, covering a wide range of topics and difficulty levels. The dataset has been instrumental in advancing research in visual question answering, promoting the development of models that integrate visual perception with language understanding.

\textbf{ImageNet} \citep{imn}: The ImageNet-1K dataset, also known as the ILSVRC 2012 dataset, is a foundational benchmark in computer vision. It comprises 1,000 distinct object categories. These categories include a diverse array of objects such as animals, vehicles, and everyday items, providing a comprehensive resource for training and evaluating image classification models.

\textbf{Flickr30k} \citep{flickr30k}: The Flickr30k dataset is a widely used benchmark for tasks involving image generation. It consists of images sourced from Flickr, each annotated with five human-generated descriptions. These annotations cover a diverse range of scenes and activities, making the dataset valuable for training and evaluating models that generate images of natural language descriptions.

\textbf{OCRVQA} \citep{oqa}: The OCR-VQA dataset is a large-scale benchmark designed to advance the field of Visual Question Answering (VQA) by focusing on questions that require reading and understanding text within images. Unlike traditional VQA datasets that emphasize object recognition and scene understanding, OCR-VQA challenges models to extract and interpret textual information from images. Questions in the dataset pertain to details such as the book's title, author, genre, and publication year, necessitating the integration of Optical Character Recognition (OCR) techniques with natural language understanding. The dataset serves as a valuable resource for developing models capable of reading and reasoning over textual content in images.

\textbf{RefCOCO} \citep{refcoco1,refcoco2}: The RefCOCO dataset is a widely used benchmark in vision-and-language research, focusing on referring expression comprehension (REC)—the task of identifying a specific object in an image based on a natural language description. Built upon images from the MS COCO dataset, RefCOCO provides annotations that link textual expressions to corresponding objects within images. RefCOCO comprises three primary subsets: RefCOCO, RefCOCO+, and RefCOCOg. The RefCOCO series has been instrumental in advancing research in referring expression comprehension and related tasks.

\textbf{HQEdit} \citep{hq}: The HQEdit dataset is a high-quality, instruction-based image editing dataset designed to advance the field of vision-language models. It comprises meticulously curated image editing instances, each consisting of an input image, a detailed textual editing instruction, and the corresponding output image. Developed using a scalable pipeline that leverages advanced foundation models such as GPT-4V and DALLE 3, HQEdit ensures high-resolution images rich in detail. The dataset includes comprehensive editing prompts and inverse-edit instructions, facilitating nuanced understanding and manipulation of visual content.

The adopted dataset statistics are as follows:

• \textbf{VQAv2 Task:} 82,783 samples for training, 5,000 samples for testing.

• \textbf{ImageNet Task:} 129,833 samples for training, 5,050 samples for testing.

• \textbf{Flickr30k Task:} 30,783 samples for training, 1,000 samples for testing.

• \textbf{OCRVQA Task:} 164,948 samples for training, 5,000 samples for testing.

• \textbf{RefCOCO Task:} 53,700 samples for training, 5,000 samples for testing.

• \textbf{HQEdit Task:} 8,813 samples for training, 1,000 samples for testing.

\begin{figure*}[htb]
\centering
\includegraphics[width=5.5in]{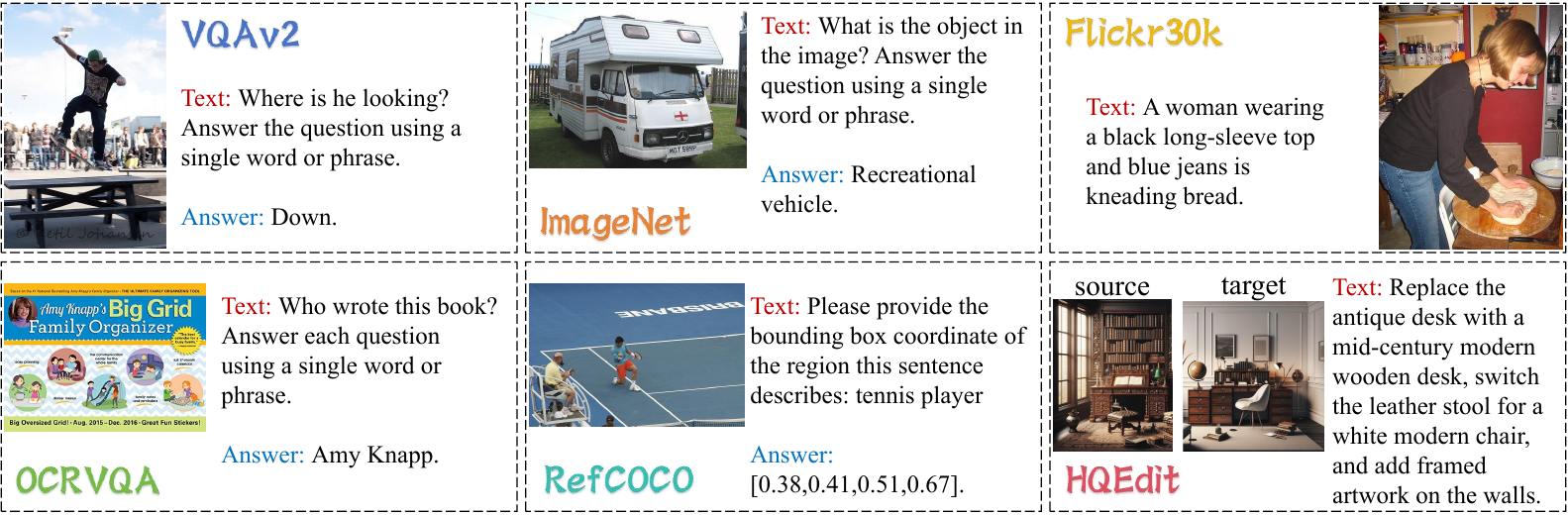}
\caption{Training and evaluation case visualization in Continual-NExT.}
\label{fig_7}
\vspace{-.3cm}
\end{figure*}

The training set of curated datasets is kept in JSONL-files, which are described as:

• \textbf{VQAv2 Task:} {"image": "xxx.jpg", "data": ["What is this photo taken looking through? Answer the question using a single word or phrase.", "net", "What position is this man playing?", "pitcher", "What color is the players shirt?", "orange", "Is this man a professional baseball player?", "yes"]}

• \textbf{ImageNet Task:} {"image": "xxx.JPEG", "data": ["What is the object in the image? Answer the question using a single word or phrase.", "Cock."]}

• \textbf{Flickr30k Task:} {"image": "xxx.jpg", "data": ["Please generate the image.", "Two young guys with shaggy hair look at their hands while hanging out in the yard."]}

• \textbf{OCRVQA Task:} {"image": "xxx.jpg", "data": ["Who wrote this book? Answer the question using a single word or phrase.", "Sandra Boynton", "What is the title of this book? ", "Mom's Family Wall Calendar 2016", "What is the genre of this book? ", "Calendars", "Is this a games related book? ", "No", "What is the year printed on this calendar? ", "2016"]}

• \textbf{RefCOCO Task:} {"image": "xxx.jpg", "data": ["Please provide the bounding box coordinate of the region this sentence describes: hot dog without peppers.", "[0.27,0.42,0.83,0.73]", "Please provide the bounding box coordinate of the region this sentence describes: hot dog without round peppers.", "[0.27,0.42,0.83,0.73]", "Please provide the bounding box coordinate of the region this sentence describes: hot dog with jalapenos.", "[0.14,0.49,0.69,0.84]", "Please provide the bounding box coordinate of the region this sentence describes: hotdog with green stuff.", "[0.14,0.49,0.69,0.84]", "Please provide the bounding box coordinate of the region this sentence describes: hotdog with pepper rings.", "[0.14,0.49,0.69,0.84]"]}

• \textbf{HQEdit Task:} {"source image": "xxx.jpg", "target image": "yyy.jpg", "instruction": "Add futuristic elements such as holographic displays and floating platforms while maintaining the market's traditional structure."}

The testing set of curated datasets is kept in JSON-files, which are described as:

• \textbf{VQAv2 Task:} {"question id": 262148000, "image": "xxx.jpg", "text": "Where is he looking? Answer the question using a single word or phrase.", "answer": "down"}

• \textbf{ImageNet Task:} {"question id": "xxx.JPEG", "image": "xxx.JPEG", "text": "What is the object in the image? Answer the question using a single word or phrase.", "answer": "Recreational vehicle"}

• \textbf{Flickr30k Task:} {"question id": "89779839", "text": "Please generate the image: A woman wearing a black long-sleeve top and blue jeans is kneading bread."}

• \textbf{OCRVQA Task:} {"question id": "1492612766 0", "image": "xxx.jpg", "text": "Who wrote this book? When the provided information is insufficient, respond with 'Unanswerable'. Answer each question using a single word or phrase.", "answer": "Amy Knapp"}

• \textbf{RefCOCO Task:} {"question id": "34136", "size": [640,424], "image": "xxx.jpg", "text": "Please provide the bounding box coordinate of the region this sentence describes: tennis player.", "origin answer bbox": "[242.94,155.68,86.22,162.53]", "answer bbox": "[0.38,0.41,0.51,0.67]"}

• \textbf{HQEdit Task:} {"source image": "xxx.jpg", "target image": "target images/yyy.jpg", "instruction": "Please generate the image: Replace the antique desk with a mid-century modern wooden desk, switch the leather stool for a white modern chair, and add framed artwork on the walls."}

To show the constructed benchmark, we plot examples in different tasks, as shown in Figure \ref{fig_7}.

\subsection{Metric Details}
\label{metricapp}
\textbf{Average Accuracy} (Avg.ACC) measures the mean test accuracy over all tasks, providing an overall assessment of the model's performance. 

\textbf{Forgetting} (FOR) quantifies the degradation in performance on previously learned tasks after the model acquires new knowledge, thus characterizing stability. 

\textbf{New Accuracy} (New.ACC) computes the average test accuracy on newly introduced tasks, serving as an indicator of the model's plasticity.

Overall, Average Accuracy, Forgetting, and New Accuracy are generally defined as:
\begin{equation}
\text{Avg.ACC} = \frac{1}{T}\sum_{i=1}^{T}A_{T,i},
\end{equation}
\begin{equation}
\text{FOR} = \frac{1}{T-1}\sum_{i=1}^{T-1}{A_{T,i} - \text{max}(A_{j,i})_{j \in [i,T-1]}},
\end{equation}
\begin{equation}
\text{New.ACC} = \frac{1}{T}\sum_{i=1}^{T}A_{i,i},
\end{equation}
where $T$ is the number of tasks, $A_{T,i}$ is the accuracy of $i$-th task on the model trained after $T$-th task, $A_{j,i}$ is the accuracy of $i$-th task on the model trained after $j$-th task, and $A_{i,i}$ is the accuracy of $i$-th task on the model trained after $i$-th task.

\textbf{Average Hallucination} (Avg.HAL) is utilized to depict hallucination-related forgetting.

\textbf{Average Instruction Unfollowing} (Avg.IUF) is designed for showing instruction unfollowing-related forgetting.

\textbf{Average Other Error} (Avg.OTH) is proposed to summarize the other errors of forgetting.

Overall, Average Hallucination, Average Instruction Unfollowing, and Average Other Error are defined as:
\begin{equation}
\text{Avg.HAL} = \frac{1}{T-1}\sum_{i=1}^{T-1}{H_{T,i} - (H_{T,i} \cap H_{i,i})},
\end{equation}
\begin{equation}
\text{Avg.IUF} =  \frac{1}{T-1}\sum_{i=1}^{T-1}{I_{T,i} - (I_{T,i} \cap I_{i,i})},
\end{equation}
\begin{equation}
\text{Avg.OTH} =  \frac{1}{T-1}\sum_{i=1}^{T-1}{O_{T,i} - (O_{T,i} \cap O_{i,i})},
\end{equation}
where $T$ is the number of tasks, $H_{T,i}$ is the hallucination of $i$-th task on the model trained after $T$-th task, $H_{i,i}$ is the hallucination of $i$-th task on the model trained after $i$-th task; $I_{T,i}$ is the instruction unfollowing of $i$-th task on the model trained after $T$-th task, $I_{i,i}$ is the instruction unfollowing of $i$-th task on the model trained after $i$-th task; $O_{T,i}$ is the other errors of $i$-th task on the model trained after $T$-th task, $O_{i,i}$ is the other errors of $i$-th task on the model trained after $i$-th task. We prompt the Qwen3-VL-30B-A3B-Instruct model to determine which type the error belongs to (prompts are shown in Figure \ref{fig:text-gene-prompt}, Figure \ref{fig:image-gene-prompt}, and Figure \ref{fig:image-edit-prompt}).

\subsection{Training Details}
\label{trainapp}
We conduct the experiments based on the official SEED-X code. The hyperparameters set in each task, excluding HQEdit, are as follows:
\begin{lstlisting}[language=Python]
    --learning_rate 1e-4 \
    --batch_size 16 \
    --weight_decay 0.05 \
    --adam_beta1 0.9 \
    --adam_beta2 0.98 \
    --adam_epsilon 1e-6 \
    --gradient_accumulation_steps 2 \
    --mixed_precision bf16 \
    --num_train_epochs 1 \
    --max_steps 3001 \
    --save_steps 3000 \
    --lr_scheduler_type cosine \
    --warmup_steps 125 \
    --min_lr_ratio 0.05 \
    --dataloader_num_workers 4 \
    --deepspeed_plugin ${PROJ_PATH}/configs/accelerate/deepspeed_stage_1.yaml
\end{lstlisting}
The hyperparameters set in HQEdit are as follows:
\begin{lstlisting}[language=Python]
    --learning_rate 1e-4 \
    --batch_size 6 \
    --weight_decay 0.05 \
    --adam_beta1 0.9 \
    --adam_beta2 0.98 \
    --adam_epsilon 1e-6 \
    --gradient_accumulation_steps 2 \
    --mixed_precision bf16 \
    --num_train_epochs 1 \
    --max_steps 10001 \
    --save_steps 10000 \
    --lr_scheduler_type cosine \
    --warmup_steps 250 \
    --min_lr_ratio 0.05 \
    --dataloader_num_workers 4 \
    --deepspeed_plugin ${PROJ_PATH}/configs/accelerate/deepspeed_stage_1.yaml
\end{lstlisting}
In addition, name of trainable parameters: connection layers and LoRAs in the adopted MLLM, SEED-X backbone, are as follows:
\begin{lstlisting}[language=Python]
patch_pos_embed
input_resampler.query
input_resampler.kv_proj.weight
input_resampler.attn.in_proj_weight
input_resampler.attn.in_proj_bias
input_resampler.attn.out_proj.weight
input_resampler.attn.out_proj.bias
input_resampler.ln_q.weight
input_resampler.ln_q.bias
input_resampler.ln_kv.weight
input_resampler.ln_kv.bias
output_resampler.query
output_resampler.kv_proj.weight
output_resampler.attn.in_proj_weight
output_resampler.attn.in_proj_bias
output_resampler.attn.out_proj.weight
output_resampler.attn.out_proj.bias
output_resampler.ln_q.weight
output_resampler.ln_q.bias
output_resampler.ln_kv.weight
output_resampler.ln_kv.bias
llm.base_model.model.model.layers.i.self_attn.rotary_emb.inv_freq
llm.base_model.model.model.embed_tokens.weight
llm.base_model.model.model.layers.i.self_attn.q_proj.lora_A.default.weight
llm.base_model.model.model.layers.i.self_attn.q_proj.lora_B.default.weight
llm.base_model.model.model.layers.i.self_attn.k_proj.lora_A.default.weight
llm.base_model.model.model.layers.i.self_attn.k_proj.lora_B.default.weight
llm.base_model.model.model.layers.i.self_attn.v_proj.lora_A.default.weight
llm.base_model.model.model.layers.i.self_attn.v_proj.lora_B.default.weight
llm.base_model.model.model.layers.i.self_attn.o_proj.lora_A.default.weight
llm.base_model.model.model.layers.i.self_attn.o_proj.lora_B.default.weight
llm.base_model.model.model.layers.i.mlp.gate_proj.lora_A.default.weight
llm.base_model.model.model.layers.i.mlp.gate_proj.lora_B.default.weight
llm.base_model.model.model.layers.i.mlp.down_proj.lora_A.default.weight
llm.base_model.model.model.layers.i.mlp.down_proj.lora_B.default.weight
llm.base_model.model.model.layers.i.mlp.up_proj.lora_A.default.weight
llm.base_model.model.model.layers.i.mlp.up_proj.lora_B.default.weight
llm.base_model.model.model.layers.i.input_layernorm.original_module.weight
llm.base_model.model.model.layers.i.input_layernorm.modules_to_save.default.weight
llm.base_model.model.model.layers.i.post_attention_layernorm.original_module.weight
llm.base_model.model.model.layers.i.post_attention_layernorm.modules_to_save.default.weight
llm.base_model.model.model.norm.original_module.weight
llm.base_model.model.model.norm.modules_to_save.default.weight
llm.base_model.model.lm_head.weight
\end{lstlisting}
(1) Setting of MoELoRA, MAGE are as follows:
\begin{lstlisting}[language=Python]
 	r: 32
 	lora_alpha: 32
 	lora_dropout: 0.05
 	expert_num: 4
\end{lstlisting}
(2) Setting of LoRA based methods are as follows:
\begin{lstlisting}[language=Python]
 	r: 32
 	lora_alpha: 32
 	lora_dropout: 0.05
\end{lstlisting}
It should be noted that all methods maintain an equal scale of trainable parameters. Although MoELoRA and MAGE introduce multi-experts mechanism, the total number of LoRA ranks remains the same as other baseline methods based on single LoRA paradigm. For instance, we set 4 experts in MoELoRA and MAGE, each LoRA rank is set to 8, maintaining the same total number of LoRA ranks as other baselines of 32.

\subsection{Baseline Details}
\label{beseapp}
\textbf{Zero-Shot} means using the SEED-X model with original pre-trained checkpoints to obtain the zero-shot performance. \textbf{LoRA Fine-Tune} \citep{lora} prepends the LoRA parameter efficient tuning paradigm into MLLM; \textbf{MoELoRA} \citep{coin} is based on the multi-expert mechanism, and the number of experts for each MoE layer is set to 4; \textbf{LAE \citep{lae}} adopts the stable EMA weight to update the LoRA parameters, which is set to 0.99. \textbf{EWC} \citep{ewc} considers the change of the training parameters and utilizes the specific parameters changing loss as a regularization penalty. \textbf{PGP} \citep{pgp} introduces a gradient projection method for efficient parameters, and changes the gradient direction orthogonal to the previous feature subspace. \textbf{CIA} \citep{cia} proposes dynamic EMA weight update strategy and instruction grouping mechanism. Considering that CIA extra incorporates the instruction grouping mechanism, which expands LoRA parameters. To make a fair comparison with other non-model expansion methods, we specifically remove the instruction grouping mechanism. \textbf{RegLoRA} \citep{sefe} identifies important LoRA parameters and applies dynamic regularization masks to limit disruptive updates, thereby stabilizing key weights to preserve prior knowledge and mitigate catastrophic forgetting.

\subsection{Training Orders}
\label{orderapp}
To validate the robustness of our method, we adopt the following three types of training orders.

1). Origin training order: VQAv2, ImageNet, Flickr30k, RefCOCO, OCRVQA, HQEdit. 

2). Reverse training order: HQEdit, OCRVQA, RefCOCO, Flickr30k, ImageNet, VQAv2. 

3). Alphabet training order: Flickr30k, HQEdit, ImageNet, OCRVQA, RefCOCO, VQAv2.

\subsection{Mathematical Deduction}
\label{deduction}
\textbf{Lemma 1:} Let $\ell(\theta; x)=\log p(x;\theta)$ be the log-likelihood for a single sample and $s(\theta;x)=\nabla_\theta \ell(\theta; x)$ be the gradient. Under standard regularity conditions that allow for the interchange of integration and differentiation, the Fisher information matrix (as shown in Eq.(\ref{ded_1})):
\begin{equation}
\label{ded_1}
I(\theta) = \mathbb{E}_{X\sim p(\cdot;\theta)}\!\left[s(\theta;X)s(\theta;X)^\top\right],
\end{equation}
is equal to the negative expected Hessian of the log-likelihood (as shown in Eq.(\ref{ded_2})):
\begin{equation}
\label{ded_2}
I(\theta) = -\mathbb{E}\big[\nabla^2_{\theta} \ell(\theta;X)\big].
\end{equation}

\textbf{Proof 1:} Since the likelihood is normalized, we have $\int p(x;\theta)\,dx = 1$. Differentiating both sides with respect to $\theta$ yields:
\begin{align}
\label{like}
0 = \nabla_\theta \int p(x;\theta)\,dx 
  = \int \nabla_\theta p(x;\theta)\,dx
  = \int p(x;\theta)\, s(\theta;x)\, dx,
\end{align}
Eq.(\ref{like}) implies that:
\begin{equation}
\mathbb{E}[s(\theta;X)] = 0.
\end{equation}
Differentiating this identity again with respect to $\theta$ and interchanging integration 
and differentiation gives that:
\begin{equation}
0 = \nabla_\theta \mathbb{E}\big[s(\theta;X)\big]
  = \int \nabla_\theta\!\left(p(x;\theta)\,s(\theta;x)\right)\!dx.
\end{equation}
Expanding the derivative, we have:
\begin{equation}
0 = \int (\nabla_\theta p)\, s\, dx + \int p\, \nabla_\theta s\, dx.
\end{equation}
Using $\nabla_\theta p(x;\theta)=p(x;\theta)s(\theta;x)$, we have:
\begin{align}
0 = \int p\, ss^\top \, dx + \int p\, \nabla_\theta s \, dx
  = \mathbb{E}[s(\theta;X)s(\theta;X)^\top] + \mathbb{E}[\nabla_\theta s(\theta;X)].
\end{align}
Since $\nabla_\theta s(\theta;x) = \nabla_\theta^2 \ell(\theta;x)$, we obtain:
\begin{equation}
I(\theta) 
= \mathbb{E}\!\left[s(\theta;X)s(\theta;X)^\top\right]
= -\mathbb{E}\!\left[\nabla^2_{\theta}\ell(\theta;X)\right].
\end{equation}

Thus, \textbf{Proof 1} holds.

The Hessian matrix \(H(\theta)=\nabla^2_\theta L(\theta)\) (or its negative) may be singular or ill-conditioned in practice, which prevents direct use of \(H(\theta)^{-1}\). A common remedy is to use a \emph{regularized inverse} of the form:
\begin{equation}
\big(H(\theta) + \lambda I\big)^{-1},\qquad \lambda>0,
\end{equation}
referred to as Tikhonov/ridge regularization or damping. The following statements make the mathematical properties of this regularization explicit.

\textbf{Proposition 1:} Let \(H\in\mathbb{R}^{d\times d}\) be a real symmetric matrix with eigen-decomposition \(H = Q \operatorname{diag}(\mu_1,\dots,\mu_d) Q^\top\) where \(Q\) is orthogonal and $\{\mu_1,\dots,\mu_d\in\mathbb{R}\}$ are the eigenvalues. For any \(\lambda\in\mathbb{R}\), the eigenvalues of \(H+\lambda I\) are \(\mu_i+\lambda\). In particular, we have the following properties:
\begin{enumerate}
  \item If \(\lambda\) is chosen so that \(\mu_i+\lambda\neq 0\) for all \(i\), then \(H+\lambda I\) is invertible and we have:
  \begin{align}
  (H+\lambda I)^{-1} = Q\,\operatorname{diag}\big((\mu_1+\lambda)^{-1},\dots, (\mu_d+\lambda)^{-1}\big)Q^\top.
  \end{align}
  \item If \(\lambda > -\min_i \{\mu_i\}\), then \(H+\lambda I\) is positive definite; in particular choosing \(\lambda>0\) guarantees invertibility whenever \(\min_i\{\mu_i\}\ge 0\).
\end{enumerate}

\textbf{Proof 2}: Considering that \(H\) is symmetric and has the orthogonal eigen decomposition \(H=Q\operatorname{diag}(\mu_i)Q^\top\), we have:
\begin{equation}
H+\lambda I = Q\operatorname{diag}(\mu_i+\lambda)Q^\top,
\end{equation}
so the eigenvalues are \((\mu_i+\lambda)\). The matrix is invertible if none of these eigenvalues vanish, which yields item 1. Item 2 follows because if \(\lambda>-\min_i\{\mu_i\}\), then \(\mu_i+\lambda>0\) for all \(i\). Hence \(H+\lambda I\) is positive definite.

\begin{figure*}[h]
\centering
\includegraphics[width=0.9\textwidth]{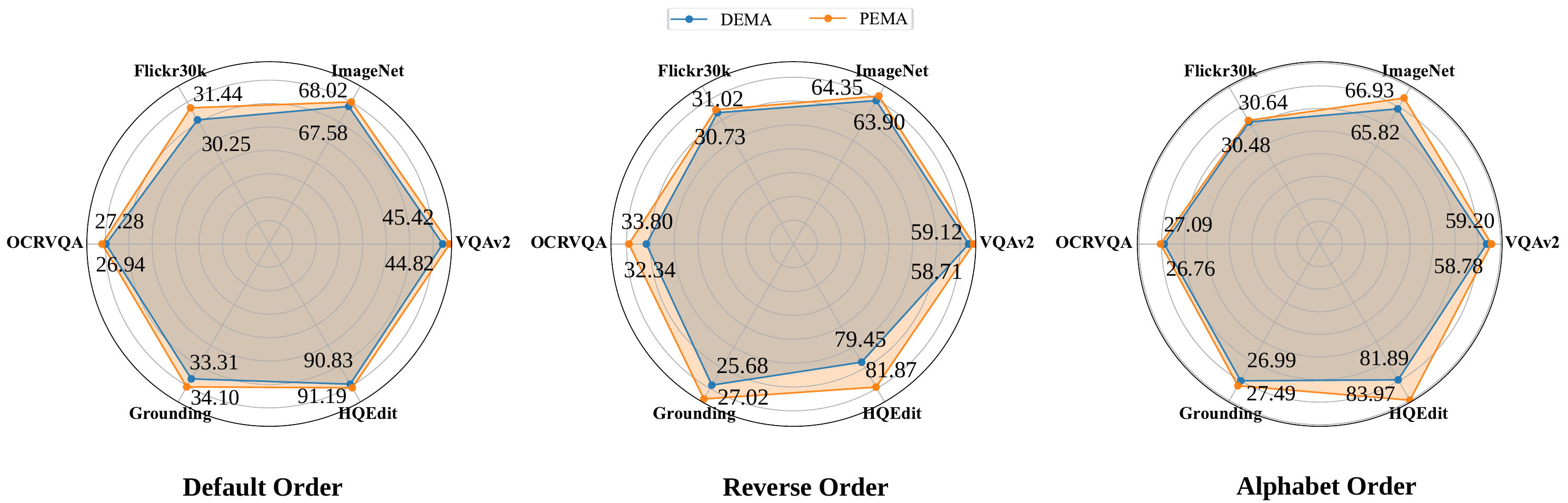}
\caption{Comparison radar chart of Accuracy on each task between PEMA and DEMA.}
\label{fig_8}
\end{figure*}
\begin{figure*}[h]
    \centering
    \begin{subfigure}[b]{0.45\textwidth}
        \centering
        \includegraphics[width=\textwidth]{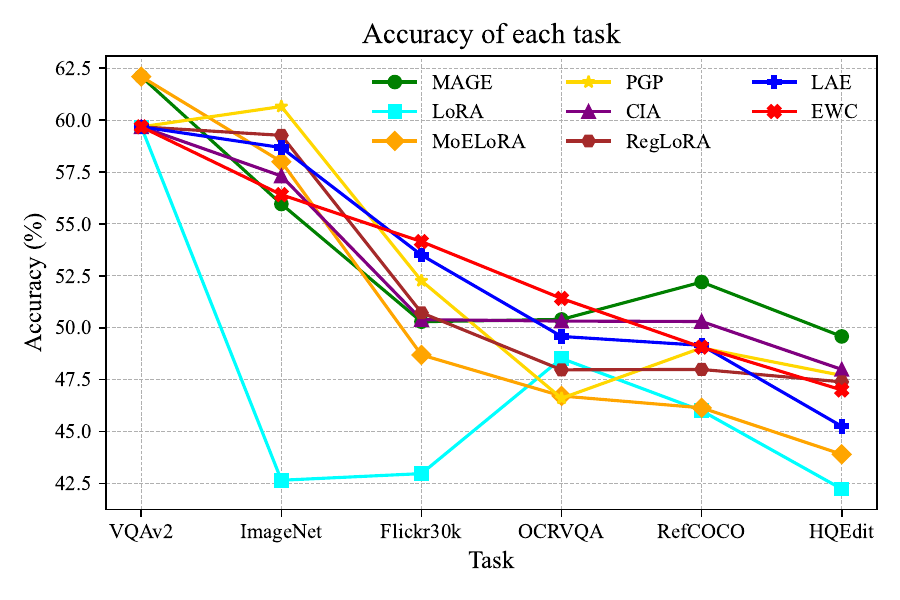}
        \caption{\textbf{Avg.ACC} performance on each task.}
        \label{fig:subfig1}
    \end{subfigure}
    \begin{subfigure}[b]{0.45\textwidth}
        \centering
        \includegraphics[width=\textwidth]{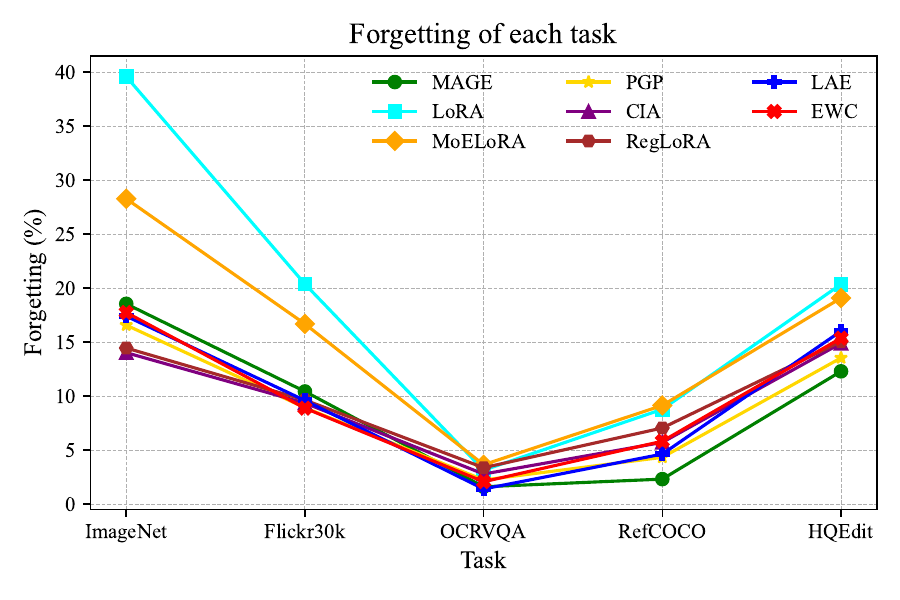}
        \caption{\textbf{Forgetting} performance on each task.}
        \label{fig:subfig2}
    \end{subfigure}
    \caption{Results of distinct methods on each task.}
    \label{fig_10}
\end{figure*}

\subsection{Distinctions and Advantages of Continual-NExT}
Compared to the previous Continual Instruction Tuning (CIT) frameworks \citep{citb,mcit,coin}, Continual-NExT has the following key distinctions and advantages: (1). Applicability to a wider range of MLLMs. CIT frameworks are limited to evaluating MLLMs that only produce text modality output (\textit{e.g.}, LLaVA \citep{llava}, QWen-VL \citep{qwen2vl}), while Continual-NExT extends support to vision-language generative MLLMs (\textit{e.g.}, SEED-X), significantly expanding its applicability. (2). Applicability to a wider range of tasks. CIT focuses solely on incremental tasks related to text generation (\textit{e.g.}, VQA), resulting in high inter-task similarity and limited diversity. In contrast, Continual-NExT additionally introduces image generation tasks (\textit{e.g.}, Image Editing). This inclusion greatly enhances task diversity and first breaks the modality boundary in continual learning by integrating multimodal generative tasks within a unified framework.

\subsection{Additional Experiments}
\label{addiexp}
To better present the results, we supplement the results of all baselines and ours training after each task (stage) and testing on previous tasks (stages) in Table \ref{maga_det}, Table \ref{lora_det}, Table \ref{moe_det}, Table \ref{ewc_det}, Table \ref{lae_det}, Table \ref{pgp_det}, Table \ref{cia_det}, and Table \ref{sefe_det}. Additionally, we visualize the Avg.ACC and Forgetting performance of different methods across stages in Figure \ref{fig_10}.

From both the Tables and the Figure, we observe that MAGE consistently outperforms the LoRA baseline at all stages, demonstrating superior continual learning capability and stronger resistance to catastrophic forgetting. Compared with other baselines, MAGE exhibits slightly inferior Avg.ACC performance in the first one or two stages, but gradually surpasses all baselines and achieves a clear performance lead in later stages.

We attribute this phenomenon to the freezing of a subset of model parameters. As a result, MAGE’s ability to adapt to newly introduced tasks is constrained. In the early stages of Unified Comprehension and Generation Continual Learning, where the number of tasks is limited, Avg.ACC is more sensitive to New.ACC. As a result, MAGE’s lower New.ACC leads to slightly weaker Avg.ACC in the initial stages. However, as the number of tasks increases, Forgetting becomes the dominant factor influencing Avg.ACC. Benefiting from its stronger anti-forgetting capability, MAGE progressively reduces performance degradation on previous tasks, allowing its Avg.ACC to steadily improve and ultimately outperform all other baselines.

We also compare the continual learning performance between MAGE with the PEMA update and MAGE with the DEMA update under three distinct training orders in Figure \ref{fig_8}. Experimental results present that our PEMA can achieve higher Accuracy in each task compared to the DEMA update with fewer storage space and computation cost.

\begin{figure*}[t]
\centering
\includegraphics[width=0.9\textwidth]{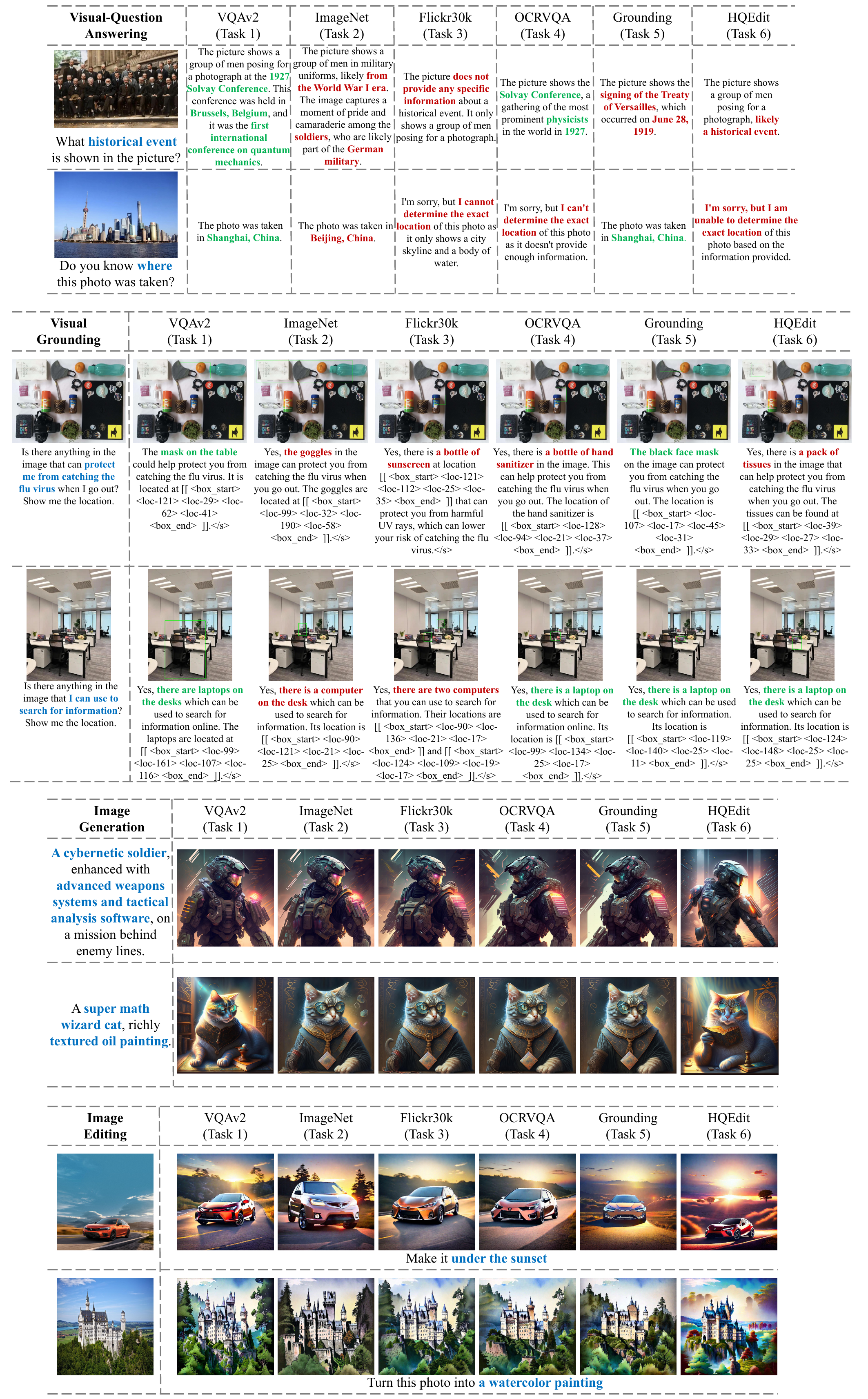}
\caption{Visualized cases after training on each task of LoRA Fine-Tune baseline.}
\label{fig_9}
\end{figure*}

We visualize the LoRA baseline’s performance on a set of comprehension and generation tests after the training of each task in Figure \ref{fig_9}. Training on certain tasks causes the model to rapidly forget previously acquired knowledge, leading to failure on some tasks. In contrast, training on other tasks facilitates positive knowledge transfer, enabling the model to recall previously learned capabilities, and thus achieving strong performance on related tasks.

The overall algorithmic pipeline of the proposed MAGE method is summarized in \textbf{Algorithm 1}.

We report the descriptive statistics about our results (\textit{e.g.}, mean and standard deviation of each metric across different methods) in Table \ref{mean_std}. Results are obtained from three runs with distinct random seeds.

\begin{figure*}[t]
\centering
\begin{tcolorbox}[
  colback=white,
  colframe=black,
  boxrule=0.8pt,
  arc=0pt,
  left=6pt,
  right=6pt,
  top=6pt,
  bottom=6pt,
  width=\textwidth
]

\textbf{Prompt (Text Generation Task):}
You are an AI response consistency evaluation agent responsible for categorizing model errors under strict definitions. You are given a question, an answer generated by a model, and its associated label. Your task is to evaluate whether the answer correctly reflects the label and follows the intended instruction in the question.

Error categories are defined as follows:

\textbf{1. Instruction Unfollowing:} This error occurs when the generated answer fails to follow the semantic intent or constraints of the intended instruction in the question. Such failures are attributed to semantic discrepancies between the instruction and the generated answer.

\textbf{2. Hallucination:} This error occurs when the generated answers wrongly introduce factually incorrect, unsupported, or illusory information that is not grounded in the given label. Hallucinated outputs typically satisfy the syntactic or structural requirements of the instruction, but significantly deviate from the ground-truth facts stated or implied by the content.

\textbf{3. Others:} Any error that does not fall into the above two categories should be classified as others.

\textbf{The question:} \{question\}
                
\textbf{The answer:} \{answer\}

\textbf{The label:} \{label\}

\textbf{Based on the above definitions, determine:}

1. Whether the generated answers follow the instruction in the question.

2. Whether hallucination is present.

3. If both are present, classify as instruction unfollowing.

\textbf{Return your decision strictly in JSON format:}

\{"error\_types": ["instruction\_not\_followed", "hallucination", "others"], "explanation": "..."\}

\end{tcolorbox}

\caption{Prompt used for text generation consistency evaluation.}
\label{fig:text-gene-prompt}

\end{figure*}

\begin{figure*}[t]
\centering
\begin{tcolorbox}[
  colback=white,
  colframe=black,
  boxrule=0.8pt,
  arc=0pt,
  left=6pt,
  right=6pt,
  top=6pt,
  bottom=6pt,
  width=\textwidth
]

\textbf{Prompt (Image Generation Task):} You are an AI image generation consistency evaluation agent responsible for categorizing model errors under strict definitions. You are given a instruction, a generated image. Your task is to evaluate whether the image correctly reflects the instruction content.

Error categories are defined as follows:

\textbf{1. Instruction Unfollowing:} This error occurs when the generated image fails to follow the semantic intent or constraints of the intended instruction. Such failures are attributed to semantic discrepancies between the instruction and the generated image.

\textbf{2. Hallucination:} This error occurs when the generated image introduces factually incorrect, unsupported, or illusory information that is not grounded in the given instruction. Hallucinated outputs typically satisfy the syntactic or structural requirements of the instruction, but significantly deviate from the ground-truth facts stated or implied by the content.

\textbf{3. Others:} Any error that does not fall into the above two categories should be classified as others.

\textbf{The instruction:} \{instruction\}

\textbf{The image:} \{image\}

\textbf{Based on the above definitions, determine:}

1. Whether the generated image follows the instruction.

2. Whether hallucination is present.

3. If both are present, classify as instruction unfollowing.

\textbf{Return your decision strictly in JSON format:}

\{"error\_types": ["instruction\_not\_followed", "hallucination", "others"], "explanation": "..."\}

\end{tcolorbox}

\caption{Prompt used for image generation consistency evaluation.}
\label{fig:image-gene-prompt}

\end{figure*}

\begin{figure*}[t]
\centering
\begin{tcolorbox}[
  colback=white,
  colframe=black,
  boxrule=0.8pt,
  arc=0pt,
  left=6pt,
  right=6pt,
  top=6pt,
  bottom=6pt,
  width=\textwidth
]

\textbf{Prompt (Image Editing Task):}
You are an AI image editing consistency evaluation agent responsible for identifying errors in edited images under strict definitions. You are given: A source image (the original image before editing). An instruction that describes how the source image should be edited. An edited image generated based on the instruction. Your task is to evaluate whether the edited image correctly reflects the editing instruction when compared to the source image.

Error categories are defined as follows:

\textbf{1. Instruction Unfollowing:} This error occurs when the edited image fails to follow the semantic intent, constraints, or required modifications described in the instruction. This includes missing required edits, incorrect changes, or changes that contradict the instruction.

\textbf{2. Hallucination:} This error occurs when the edited image introduces new visual elements, attributes, or changes that are not requested or implied by the instruction. The image may appear realistic, but the added content is unsupported by the instruction.

\textbf{3. Others:} Any error that does not fall into the above two categories, including minor artifacts or ambiguities that cannot be clearly attributed to instruction unfollowing or hallucination.

\textbf{The Source Image:} \{source\_image\}

\textbf{The Instruction:} \{instruction\}

\textbf{The Edited Image:} \{edited\_image\}

\textbf{Based on the above definitions, determine:}

1. Whether the generated image follows the instruction.

2. Whether hallucination is present.

3. If both are present, classify as instruction unfollowing.

\textbf{Return your decision strictly in JSON format:}

\{"error\_types": "instruction\_not\_followed", "hallucination", "others", "explanation": "..."\}

\end{tcolorbox}

\caption{Prompt used for image editing consistency evaluation.}
\label{fig:image-edit-prompt}

\end{figure*}

\begin{table*}[th]
\vspace{-3mm}
\renewcommand{\arraystretch}{2.0}
\caption{Comparison of Avg. ACC, Forgetting, and New ACC between our method and the baselines. The accuracy for each task is measured after training on the final task. We report the mean and standard deviation of each metric across different methods as $\text{mean}\pm\text{std}$. Results are obtained from three runs with distinct random seeds. \looseness=-1}
\label{mean_std}
\centering
\resizebox{1.0\linewidth}{!}{
\begin{tabular}{l|ccccccccc}
\toprule
\textbf{Metrics} & \textbf{LoRA} & \textbf{MoELoRA} & \textbf{EWC} & \textbf{LAE} & \textbf{PGP} & \textbf{CIA} & \textbf{RegLoRA} & \textbf{Ours(MAGE)} & \textbf{Upper-Bound} \\
\midrule
\textbf{Avg.ACC} & 42.23$\pm${0.83} & 43.90$\pm${0.47} & 47.00$\pm${0.85} & 45.25$\pm${0.61} & 47.70$\pm${0.52} & 47.99$\pm${0.18} & 47.39$\pm${0.67} & 49.58$\pm${0.58} & 53.31$\pm${0.49} \\
\textbf{Forgetting} & 20.36$\pm${0.31} & 19.08$\pm${0.22} & 15.85$\pm${0.19} & 16.29$\pm${0.28} & 13.54$\pm${0.37} & 14.89$\pm${0.16} & 15.05$\pm${0.49} & 12.26$\pm${0.13} & - \\
\textbf{New.ACC} & 59.19$\pm${0.73} & 59.80$\pm${0.50} & 60.21$\pm${0.75} & 58.82$\pm${0.38} & 58.97$\pm${0.76} & 60.40$\pm${0.55} & 59.94$\pm${0.42} & 59.79$\pm${0.63} & - \\
\bottomrule
\end{tabular}}
\vspace{-3mm}
\end{table*}

\begin{table*}[th]
\renewcommand{\arraystretch}{2.0}
\caption{Detailed raw performance after training on each task of our proposed \textit{MAGE}.}
\label{maga_det}
\centering
\resizebox{0.65\linewidth}{!}{
\begin{tabular}{l|cccccc|c}
\toprule
\multirow{2}{*}{\textbf{Method}} & \multicolumn{6}{c|}{\textbf{Accuracy on Each Task}} & \multirow{2}{*}{\textbf{\textcolor[rgb]{0.773,0.353,0.067}{Avg.ACC($\uparrow$)}}} \\ \cline{2-7} 
& \multicolumn{1}{c}{\textcolor{someblue}{VQAv2}} & \multicolumn{1}{c}{\textcolor{someblue}{ImageNet}} & \multicolumn{1}{c}{\textcolor{someblue}{Flickr30k}} & \multicolumn{1}{c}{\textcolor{someblue}{OCRVQA}} & \multicolumn{1}{c}{\textcolor{someblue}{RefCOCO}} & \multicolumn{1}{c|}{\textcolor{someblue}{HQEdit}} \\ \hline
\multirow{6}{*}{\textbf{MAGE}} & 62.10 & - & - & - & - & - & 62.10 \\
 & 43.56 & 68.35 & -     & -     & -     & -     & 55.96 \\
 & 41.25 & 78.37 & 31.26 & -     & -     & -     & 50.29 \\
 & 57.29 & 75.18 & 32.02 & 37.09 & -     & -     & 50.40 \\
 & 55.84 & 70.56 & 31.75 & 34.10 & 68.74 & -     & 52.20 \\ 
 & 45.42 & 68.02 & 31.44 & 27.28 & 34.10 & 91.19 & 49.58 \\
\bottomrule
\end{tabular}}
\vspace{-3mm}
\end{table*}

\begin{table*}[th]
\renewcommand{\arraystretch}{2.0}
\caption{Detailed raw performance after training on each task of \textit{LoRA} baseline.}
\label{lora_det}
\centering
\resizebox{0.65\linewidth}{!}{
\begin{tabular}{l|cccccc|c}
\toprule
\multirow{2}{*}{\textbf{Method}} & \multicolumn{6}{c|}{\textbf{Accuracy on Each Task}} & \multirow{2}{*}{\textbf{\textcolor[rgb]{0.773,0.353,0.067}{Avg.ACC($\uparrow$)}}} \\ \cline{2-7} 
& \multicolumn{1}{c}{\textcolor{someblue}{VQAv2}} & \multicolumn{1}{c}{\textcolor{someblue}{ImageNet}} & \multicolumn{1}{c}{\textcolor{someblue}{Flickr30k}} & \multicolumn{1}{c}{\textcolor{someblue}{OCRVQA}} & \multicolumn{1}{c}{\textcolor{someblue}{RefCOCO}} & \multicolumn{1}{c|}{\textcolor{someblue}{HQEdit}} \\ \hline
\multirow{6}{*}{\textbf{LoRA}} & 59.68 & - & - & - & - & - & 59.68 \\
 & 20.10 & 65.19 & -     & -     & -     & -     & 42.65 \\
 & 18.94 & 78.80 & 31.18 & -     & -     & -     & 42.97 \\
 & 50.26 & 71.32 & 31.26 & 41.22 & -     & -     & 48.52 \\
 & 36.75 & 66.43 & 30.94 & 29.28 & 66.60 & -     & 46.00 \\ 
 & 18.40 & 61.01 & 30.60 & 19.38 & 32.70 & 91.27 & 42.23 \\
\bottomrule
\end{tabular}}
\vspace{-3mm}
\end{table*}

\begin{table*}[th]
\renewcommand{\arraystretch}{2.0}
\caption{Detailed raw performance after training on each task of \textit{MoELoRA} baseline.}
\label{moe_det}
\centering
\resizebox{0.65\linewidth}{!}{
\begin{tabular}{l|cccccc|c}
\toprule
\multirow{2}{*}{\textbf{Method}} & \multicolumn{6}{c|}{\textbf{Accuracy on Each Task}} & \multirow{2}{*}{\textbf{\textcolor[rgb]{0.773,0.353,0.067}{Avg.ACC($\uparrow$)}}} \\ \cline{2-7} 
& \multicolumn{1}{c}{\textcolor{someblue}{VQAv2}} & \multicolumn{1}{c}{\textcolor{someblue}{ImageNet}} & \multicolumn{1}{c}{\textcolor{someblue}{Flickr30k}} & \multicolumn{1}{c}{\textcolor{someblue}{OCRVQA}} & \multicolumn{1}{c}{\textcolor{someblue}{RefCOCO}} & \multicolumn{1}{c|}{\textcolor{someblue}{HQEdit}} \\ \hline
\multirow{6}{*}{\textbf{MoELoRA}} & 62.10 & - & - & - & - & - & 62.10 \\
 & 33.83 & 82.16 & -     & -     & -     & -     & 58.00 \\
 & 28.74 & 85.86 & 31.46 & -     & -     & -     & 48.69 \\
 & 56.37 & 76.95 & 31.99 & 21.52 & -     & -     & 46.71 \\
 & 42.81 & 68.53 & 31.21 & 18.15 & 70.00 & -     & 46.14 \\ 
 & 25.32 & 63.57 & 30.59 & 16.04 & 36.32 & 91.54 & 43.90 \\
\bottomrule
\end{tabular}}
\vspace{-3mm}
\end{table*}

\begin{table*}[th]
\renewcommand{\arraystretch}{2.0}
\caption{Detailed raw performance after training on each task of \textit{EWC} baseline.}
\label{ewc_det}
\centering
\resizebox{0.65\linewidth}{!}{
\begin{tabular}{l|cccccc|c}
\toprule
\multirow{2}{*}{\textbf{Method}} & \multicolumn{6}{c|}{\textbf{Accuracy on Each Task}} & \multirow{2}{*}{\textbf{\textcolor[rgb]{0.773,0.353,0.067}{Avg.ACC($\uparrow$)}}} \\ \cline{2-7} 
& \multicolumn{1}{c}{\textcolor{someblue}{VQAv2}} & \multicolumn{1}{c}{\textcolor{someblue}{ImageNet}} & \multicolumn{1}{c}{\textcolor{someblue}{Flickr30k}} & \multicolumn{1}{c}{\textcolor{someblue}{OCRVQA}} & \multicolumn{1}{c}{\textcolor{someblue}{RefCOCO}} & \multicolumn{1}{c|}{\textcolor{someblue}{HQEdit}} \\ \hline
\multirow{6}{*}{\textbf{EWC}} & 59.68 & - & - & - & - & - & 59.68 \\
 & 41.94 & 73.29 & -     & -     & -     & -     & 57.62 \\
 & 41.92 & 89.41 & 31.11 & -     & -     & -     & 54.15 \\
 & 53.46 & 82.48 & 32.21 & 37.50 & -     & -     & 51.41 \\
 & 47.98 & 67.31 & 32.15 & 29.47 & 68.40 & -     & 49.06 \\ 
 & 40.44 & 61.07 & 30.49 & 18.64 & 40.08 & 91.30 & 47.00 \\
\bottomrule
\end{tabular}}
\vspace{-3mm}
\end{table*}

\begin{table*}[th]
\renewcommand{\arraystretch}{2.0}
\caption{Detailed raw performance after training on each task of \textit{LAE} baseline.}
\label{lae_det}
\centering
\resizebox{0.65\linewidth}{!}{
\begin{tabular}{l|cccccc|c}
\toprule
\multirow{2}{*}{\textbf{Method}} & \multicolumn{6}{c|}{\textbf{Accuracy on Each Task}} & \multirow{2}{*}{\textbf{\textcolor[rgb]{0.773,0.353,0.067}{Avg.ACC($\uparrow$)}}} \\ \cline{2-7} 
& \multicolumn{1}{c}{\textcolor{someblue}{VQAv2}} & \multicolumn{1}{c}{\textcolor{someblue}{ImageNet}} & \multicolumn{1}{c}{\textcolor{someblue}{Flickr30k}} & \multicolumn{1}{c}{\textcolor{someblue}{OCRVQA}} & \multicolumn{1}{c}{\textcolor{someblue}{RefCOCO}} & \multicolumn{1}{c|}{\textcolor{someblue}{HQEdit}} \\ \hline
\multirow{6}{*}{\textbf{LAE}} & 59.68 & - & - & - & - & - & 59.68 \\
 & 42.28 & 77.49 & -     & -     & -     & -     & 59.89 \\
 & 40.50 & 88.81 & 31.18 & -     & -     & -     & 53.50 \\
 & 55.50 & 83.29 & 32.04 & 27.48 & -     & -     & 49.58 \\
 & 42.36 & 79.53 & 31.87 & 26.28 & 65.72 & -     & 49.15 \\ 
 & 21.56 & 70.71 & 30.63 & 28.54 & 28.68 & 91.39 & 45.25 \\
\bottomrule
\end{tabular}}
\vspace{-3mm}
\end{table*}

\begin{table*}[th]
\renewcommand{\arraystretch}{2.0}
\caption{Detailed raw performance after training on each task of \textit{PGP} baseline.}
\label{pgp_det}
\centering
\resizebox{0.65\linewidth}{!}{
\begin{tabular}{l|cccccc|c}
\toprule
\multirow{2}{*}{\textbf{Method}} & \multicolumn{6}{c|}{\textbf{Accuracy on Each Task}} & \multirow{2}{*}{\textbf{\textcolor[rgb]{0.773,0.353,0.067}{Avg.ACC($\uparrow$)}}} \\ \cline{2-7} 
& \multicolumn{1}{c}{\textcolor{someblue}{VQAv2}} & \multicolumn{1}{c}{\textcolor{someblue}{ImageNet}} & \multicolumn{1}{c}{\textcolor{someblue}{Flickr30k}} & \multicolumn{1}{c}{\textcolor{someblue}{OCRVQA}} & \multicolumn{1}{c}{\textcolor{someblue}{RefCOCO}} & \multicolumn{1}{c|}{\textcolor{someblue}{HQEdit}} \\ \hline
\multirow{6}{*}{\textbf{PGP}} & 59.68 & - & - & - & - & - & 59.68 \\
 & 43.11 & 78.20 & -     & -     & -     & -     & 60.66 \\
 & 41.83 & 83.56 & 31.40 & -     & -     & -     & 52.26 \\
 & 52.98 & 81.59 & 31.88 & 19.96 & -     & -     & 46.60 \\
 & 45.45 & 77.76 & 30.32 & 18.40 & 73.20 & -     & 49.03 \\ 
 & 38.08 & 74.14 & 30.82 & 18.08 & 33.64 & 91.43 & 47.70 \\
\bottomrule
\end{tabular}}
\vspace{-3mm}
\end{table*}

\begin{table*}[th]
\renewcommand{\arraystretch}{2.0}
\caption{Detailed raw performance after training on each task of \textit{CIA} baseline.}
\label{cia_det}
\centering
\resizebox{0.65\linewidth}{!}{
\begin{tabular}{l|cccccc|c}
\toprule
\multirow{2}{*}{\textbf{Method}} & \multicolumn{6}{c|}{\textbf{Accuracy on Each Task}} & \multirow{2}{*}{\textbf{\textcolor[rgb]{0.773,0.353,0.067}{Avg.ACC($\uparrow$)}}} \\ \cline{2-7} 
& \multicolumn{1}{c}{\textcolor{someblue}{VQAv2}} & \multicolumn{1}{c}{\textcolor{someblue}{ImageNet}} & \multicolumn{1}{c}{\textcolor{someblue}{Flickr30k}} & \multicolumn{1}{c}{\textcolor{someblue}{OCRVQA}} & \multicolumn{1}{c}{\textcolor{someblue}{RefCOCO}} & \multicolumn{1}{c|}{\textcolor{someblue}{HQEdit}} \\ \hline
\multirow{6}{*}{\textbf{CIA}} & 59.68 & - & - & - & - & - & 59.68 \\
 & 45.66 & 68.95 & -     & -     & -     & -     & 57.31 \\
 & 40.98 & 79.12 & 31.08 & -     & -     & -     & 50.39 \\
 & 51.33 & 77.38 & 31.50 & 41.06 & -     & -     & 50.32 \\
 & 42.79 & 72.27 & 30.89 & 35.29 & 70.28 & -     & 50.30 \\ 
 & 28.40 & 68.91 & 30.49 & 31.00 & 37.78 & 91.36 & 47.99 \\
\bottomrule
\end{tabular}}
\vspace{-3mm}
\end{table*}

\begin{table*}[th]
\renewcommand{\arraystretch}{2.0}
\caption{Detailed raw performance after training on each task of \textit{RegLoRA} baseline.}
\label{sefe_det}
\centering
\resizebox{0.65\linewidth}{!}{
\begin{tabular}{l|cccccc|c}
\toprule
\multirow{2}{*}{\textbf{Method}} & \multicolumn{6}{c|}{\textbf{Accuracy on Each Task}} & \multirow{2}{*}{\textbf{\textcolor[rgb]{0.773,0.353,0.067}{Avg.ACC($\uparrow$)}}} \\ \cline{2-7} 
& \multicolumn{1}{c}{\textcolor{someblue}{VQAv2}} & \multicolumn{1}{c}{\textcolor{someblue}{ImageNet}} & \multicolumn{1}{c}{\textcolor{someblue}{Flickr30k}} & \multicolumn{1}{c}{\textcolor{someblue}{OCRVQA}} & \multicolumn{1}{c}{\textcolor{someblue}{RefCOCO}} & \multicolumn{1}{c|}{\textcolor{someblue}{HQEdit}} \\ \hline
\multirow{6}{*}{\textbf{RegLoRA}} & 59.68 & - & - & - & - & - & 59.68 \\
 & 45.23 & 73.31 & -     & -     & -     & -     & 59.27 \\
 & 40.41 & 80.47 & 31.29 & -     & -     & -     & 50.72 \\
 & 52.17 & 70.72 & 32.06 & 36.92 & -     & -     & 47.97 \\
 & 45.43 & 65.61 & 31.39 & 30.68 & 66.82 & -     & 47.99 \\ 
 & 40.34 & 61.10 & 30.42 & 25.72 & 35.18 & 91.60 & 47.39 \\
\bottomrule
\end{tabular}}
\vspace{-3mm}
\end{table*}

\begin{algorithm*}
        \caption{Training phase of MAGE}
        \KwIn{Pre-trained ViT model $f_{vis}$, Pre-trained Large Language Model $f_{lan}$ with inserted General LoRA $f_{GI}$ and $f_{GT}$, Expert LoRA $f_{EI}$ and $f_{ET}$, connection layers $f_{con}$, embedding layer $\phi$, number of tasks $D$, number of iterations $T$, training set ${\{\{{I}_i^t, {T}_i^t, {Y}_i^t\}_{i=1}^{n_t}\}}_{t=1}^T$, learning rate $\eta$, loss function $\mathcal{L}_x$.}
        \KwOut{$f_{GI}^*$, $f_{GT}^*$, $f_{EI}^*$, $f_{ET}^*$ ($^*$ denotes the EMA parameters), and $f_{con}$.}
        \textbf{Initialize:} $f_{GI}$, $f_{GT}$, $f_{EI}$, $f_{ET}$, $f_{con}$. \\
        \For {$d$ = 1, ..., $D$}
        {
        1. Freeze the parts of LoRA unrelated to the current task. \\
        \For {$epoch$ = 1}
        {   
            \For{$t$ = 1, ..., $T$}
            {
                2.Draw a mini-batch $B$ = $({I}_i^t, {T}_i^t, {Y}_i^t\}_{i=1}^{n_t}$. \\
                \For{$({I}, {T}, {Y})$ in $B$}
                {   
                    3.Encode $I$ into $I_t$ by $I_t = f_{vis}(I).$ \\
                    4.Embed $T$ into $T_t$ by $T_t = \phi(T).$ \\
                    5.Prepend $I_t$ with $T_t$ by $[T_t; I_t]$. \\
                    6.Obtain autoregressive prediction by $\hat{Y}=f_{lan}([T_t; I_t])$. \\
                    7.Calculate loss $\mathcal{L}_B$ by accumulating $\mathcal{L}(Y,\hat{Y})$. \\
                    8.Backward propagation and update trainable parameters with optimizer. \\
                    \# Fisher information matrix calculate. \\
                    9.Calculate Fisher matrix according to Eq.(\ref{fisher}). \\
                    \# EMA weight calculate.  \\
                    10.Obtain regularized Hessian matrix according to Eq.(\ref{regrev}). \\
                    11.Calculate $\beta_t$ according to Eq.(\ref{weight}). \\
                    \# EMA parameter update. \\
                    12.Update EMA parameters with parameter-wise $\beta_t$. \\
                }
            }
        }
        13. Save checkpoints of $f_{GI}^*$, $f_{GT}^*$, $f_{EI}^*$, $f_{ET}^*$, and $f_{con}$. \\
        }
\end{algorithm*}

\end{document}